\documentclass{article} 
\usepackage{iclr2021_conference,times}

\usepackage{amsmath,amsfonts,bm}

\def\eqref#1{equation~\ref{#1}}

\def\1{\bm{1}}

\DeclareMathAlphabet{\mathsfit}{\encodingdefault}{\sfdefault}{m}{sl}
\SetMathAlphabet{\mathsfit}{bold}{\encodingdefault}{\sfdefault}{bx}{n}

\DeclareMathOperator*{\argmax}{arg\,max}

\usepackage{hyperref}
\usepackage{url}

\usepackage{graphicx}
\graphicspath{{Figures/}}

\usepackage{listings}
\usepackage{xcolor,colortbl}

\definecolor{Gray}{gray}{0.85}
\definecolor{LightCyan}{rgb}{0.48,0.78,1}
\definecolor{LightColor1}{rgb}{0.68,0.88,1.2}

\definecolor{LightGray}{gray}{.8}
\newcolumntype{C}[1]{>{\centering\arraybackslash}p{#1}}

\newcolumntype{y}{>{\columncolor{yellow}}c}
\newcolumntype{a}{>{\columncolor{lightgray}}c}
\newcolumntype{g}{>{\columncolor{pink}}c}
\newcolumntype{o}{>{\columncolor{LightColor1}}c}
\newcolumntype{u}{>{\columncolor{LightCyan}}c}

\newcommand{\specialcell}[2][c]{%
	\begin{tabular}[#1]{@{}c@{}}#2\end{tabular}}

\PassOptionsToPackage{hyphens}{url}
\usepackage[T1]{fontenc}

\usepackage{amsmath,amssymb,amsfonts} 
\usepackage{fixltx2e}
\usepackage{multirow}
\usepackage{booktabs}
\usepackage{amsmath}

\usepackage{subcaption}

\usepackage{makecell}
\usepackage[T1]{fontenc}

\lstdefinestyle{customc}{
	belowcaptionskip=1\baselineskip,
	breaklines=true,
	frame=L,
	xleftmargin=\parindent,
	language=C,
	showstringspaces=false,
	basicstyle=\footnotesize\ttfamily,
	keywordstyle=\bfseries\color{green!40!black},
	commentstyle=\itshape\color{purple!40!black},
	identifierstyle=\color{blue},
	stringstyle=\color{orange},
}

\lstdefinestyle{customasm}{
	belowcaptionskip=1\baselineskip,
	frame=L,
	xleftmargin=\parindent,
	language=[x86masm]Assembler,
	basicstyle=\footnotesize\ttfamily,
	commentstyle=\itshape\color{purple!40!black},
}

\hypersetup{
	colorlinks=true,
	linkcolor=blue,
	filecolor=green,      
	urlcolor=blue,
	citecolor=black
}
\usepackage{soul}
\definecolor{mycolor}{rgb}{0.75, 0.75, 0.75}
\definecolor{lavendergray}{rgb}{0.77, 0.76, 0.82}
\definecolor{lightgray}{rgb}{0.83, 0.83, 0.83}

\title{
	Derivative Manipulation for \\General Example Weighting 
}

\author{%
	Xinshao Wang\thanks{
		Work mainly done at Queen's University Belfast and Anyvision.
	}
	\\
	University of Oxford
	\\ 
	\texttt{xinshao.wang at eng.ox.ac.uk} \\
	\And
	Elyor Kodirov \\
	Anyvision Research Team \\
	\texttt{elyor at anyvision.co} \\
	\And
	Yang Hua \\
	Queen's University Belfast\\
	\texttt{y.hua at qub.ac.uk} \\
	\And
	~~~~~~~~~~~~~~~~~~~~~~~~~~~~~~~~~~~~~~~~~~~~~~~~~~~~~~~~Neil M. Robertson \\
	~~~~~~~~~~~~~~~~~~~~~~~~~~~~~~~~~~~~~~~~~~~~~~~~~~~~~~~~Queen's University Belfast \\
	~~~~~~~~~~~~~~~~~~~~~~~~~~~~~~~~~~~~~~~~~~~~~~~~~~~~~~~~\texttt{n.robertson at qub.ac.uk} 
}

\iclrfinalcopy 
\begin{document}

\maketitle

\vspace{-0.4cm}
\begin{abstract}
	Real-world large-scale datasets usually contain noisy labels and are imbalanced.
	Therefore, we propose derivative manipulation (DM), a novel and general example weighting approach for training robust deep models under these adverse conditions.
	%
	DM has two main merits.  
	First, loss function and example weighting are common techniques in the literature. 
	DM reveals their connection (a loss function does example weighting) and is a replacement of both.  
	%
	Second, 
	despite that a loss defines an example weighting scheme by its derivative, in the loss design, we need to consider whether it is differentiable. 
	Instead, DM is more flexible by directly modifying the derivative so that a loss can be a non-elementary format too.     
	%
	%
	%
	%
	Technically, DM defines an emphasis density function by a derivative magnitude function.
	DM is generic in that diverse weighting schemes can be derived.
	%
	Extensive experiments on both vision and language tasks prove DM's effectiveness.
\end{abstract}

\section{Introduction}
\label{introduction}
\vspace{-0.1cm}


In large-scale deep learning tasks, addressing label noise and sample imbalance is fundamental and has been widely studied \citep{reed2015training,chang2017active,wang2019imae}. There are two popular approaches, i.e., loss design \citep{ghosh2017robust,zhang2018generalized,wang2019imae} and example weighting design \citep{chang2017active,ren2018learning,shu2019meta,jiang2015self,jiang2018mentornet,jiang2020beyond}, due to their easy implementation and widely demonstrated effectiveness. 
We reveal that a loss function does example weighting and propose DM to replace them.  
%
On the one hand, the derivative magnitude of an example decides how much impact it has on updating a model \citep{hampel1986approach,barron2019general}. Hence, a derivative magnitude function defines a weighting scheme over training examples. 
On the other hand, in gradient-based optimisation, the role of a loss function is to provide gradient used for back-propagation. 
DM designs the derivative directly so that we do not need to derive it from a loss function. 
%

\vspace{-0.1cm}
\subsection{The relationship between derivative magnitude and loss value}
\label{sec:incompatibility}
\vspace{-0.1cm}

We present two partially incompatible perspectives on the robustness of a loss function, which motivates us to design the derivative other than a loss function: 
(1) \textit{Loss value.} From this viewpoint, a loss function, which is less sensitive to large errors (i.e., residuals), is more robust and preferred \citep{hastie2015statistical,huber1981robust}.
For example, absolute error is considered more robust than the squared error.  
An outlier has a larger error by definition, but its loss value should not increase dramatically when a robust loss function is applied. 
(2) \textit{Derivative magnitude}\footnote{Throughout this paper, derivative refers to the derivative with respect to logits instead of probabilities, inspired by \citep{wang2019imae}. We propose to design it, instead of deriving it from a loss, thus being special.}. A more robust model is less affected by noisy data than clean data. 
Therefore, a noisy example should have a smaller derivative magnitude.   
However, as proved by \citep{janocha2016loss} and our mathematical analysis illustrated in Figure~\ref{fig:DifferentWeighting_MAE_MSE_CCE_GCE_DM}, \textit{loss functions may have non-monotonic partial derivatives}. 
In such cases, \textit{two viewpoints are not entirely consistent and the first one becomes misleading when inconsistency occurs}. 
We support the second because the role of a loss is offering the gradient to back-propagate as shown in Figure~\ref{fig:Illustration_DM_Loss}. 
Therefore, derivative magnitude is a direct angle, thus being reliable.  
%
%

Although it may be feasible to design a loss function which offers the desired derivative, we need to consider whether it is differentiable and derive its derivative to get its underlying weighting scheme. 
This ``two-step'' procedure makes the design of a loss more complicated than DM.


\vspace{-0.1cm}
\subsection{Defining example-level weighting schemes by emphasis density functions}
\vspace{-0.1cm}

\begin{figure}[!t]
	\centering
	\begin{subfigure}[h!]{0.53\textwidth}
		\vspace{0.8cm}
		\centering
		\captionsetup{width=1.04\textwidth}
		\includegraphics[clip, trim=-0.35cm 0.1cm 0.0cm 0.4cm, width=0.85\textwidth]{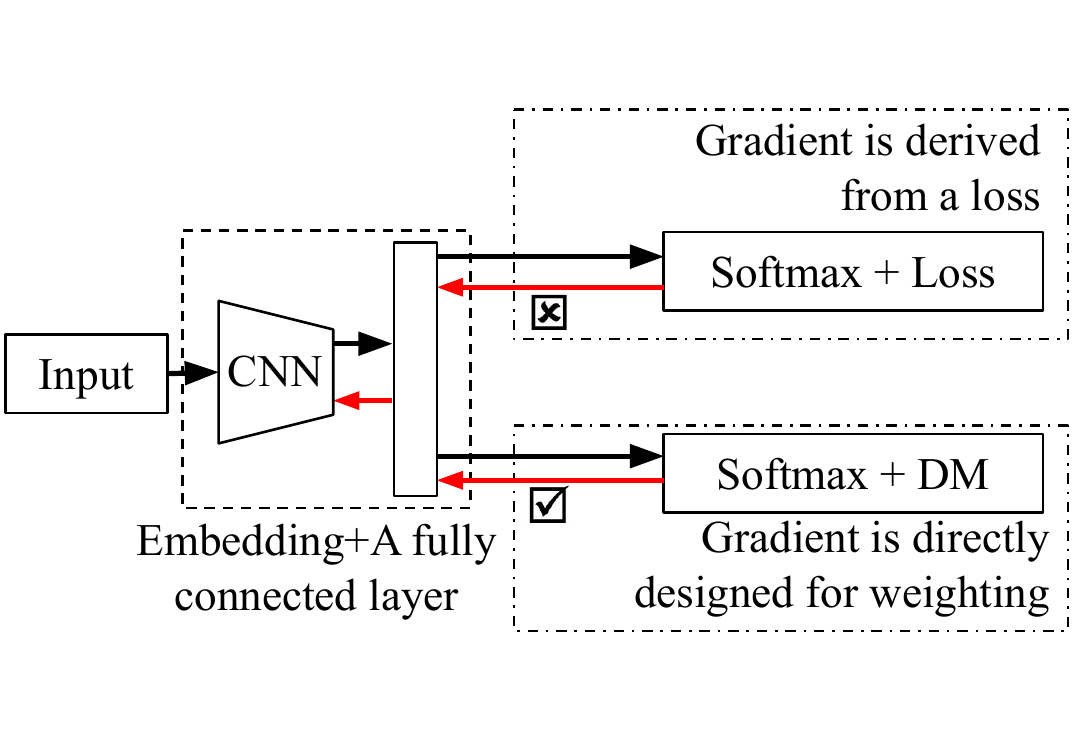}
		\caption{ 
			Common practice and DM. Black and red arrows denote forward process and gradient back-propagation, respectively.    
		}
		\label{fig:Illustration_DM_Loss}
		\vspace{0.05cm}
	\end{subfigure}
	\hfil
	\begin{subfigure}[h!]{0.45\textwidth}
		\vspace{-0.2cm}
		\centering
		\captionsetup{width=0.910\textwidth}
		\includegraphics[clip, trim=3.4cm 9.08cm 4.1cm 9.22cm, width=0.95\textwidth]{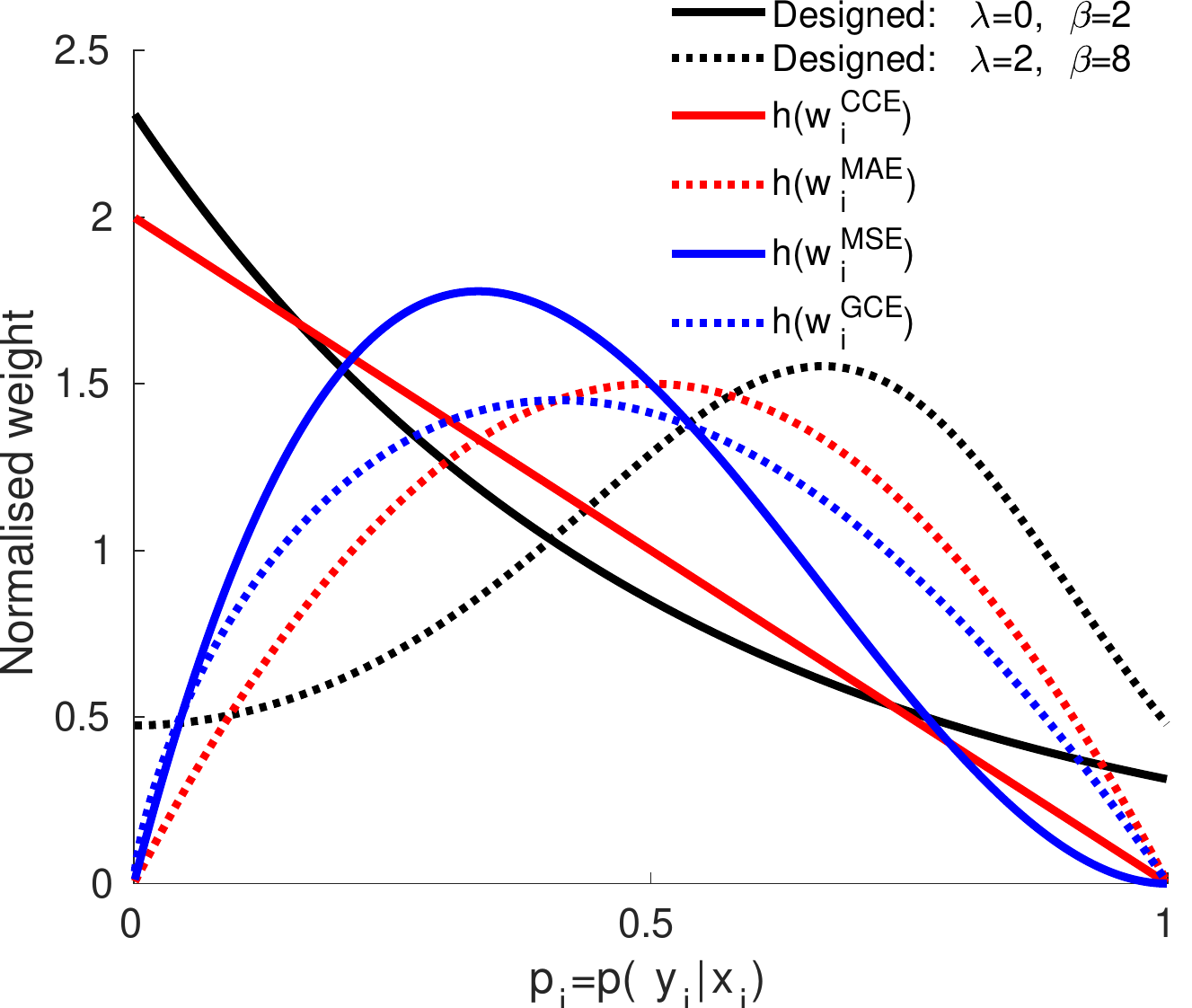}
		\caption{ 
			An EDF is an example-level weighting function normalised by its integral over $[0,1]$.    
		}
		\label{fig:DifferentWeighting_MAE_MSE_CCE_GCE_DM}
		\vspace{0.10cm}
	\end{subfigure}
	%
	%
	\vspace{-0.3cm}
	\caption{
		Illustration of DM in terms of optimisation and example weighting.
	}
	\label{fig:DM_Figure_Illustration}
	\vspace{-0.42cm}
\end{figure}

DM nonlinearly transforms the derivative magnitude, followed by derivative normalisation (DN) so that the total emphasis (weight) is one unit. 
We term a normalised derivative magnitude function an Emphasis Density Function (EDF), which explicitly defines an example-level weighting scheme over data points. 
An EDF considers emphasis mode and variance, being analogous to a probability density function (PDF), thus we can design an EDF according to existing PDFs. 
\textit{Emphasis mode} represents examples whose weight values are the largest. 
%
\textit{Emphasis variance} decides the spread of emphasis (weight), i.e., the variance of an EDF curve. 
We show some EDFs in Figure~\ref{fig:DifferentWeighting_MAE_MSE_CCE_GCE_DM}. 
Intuitively, \textit{examples of greater ``interest'' should contribute more to a model's update}. 
Emphasis variance controls how significantly they are emphasised. 
For example, depending on a scenario, it can be defined to focus on easy, semi-hard, or hard examples.
Therefore, conceptually, DM is a simplified generic form of many existing heuristically-designed example weighting methods \citep{li2017learning,malach2017decoupling,ren2018learning,han2018co,jiang2015self,jiang2018mentornet,jiang2020beyond}.

%

We summarise our contributions: 
(1) We propose a novel approach that unifies the design of example weighting
and loss function in a single framework, named DM. (2) We demonstrate the effectiveness of DM for training robust
deep networks on diverse tasks: (a) Image classification under synthetic and real-world label noise;
(b) Video retrieval with unknown and diverse abnormal examples; (c) Sentiment classification of
movie reviews when label noise and sample imbalance exist. (3) We show DM with diverse network
architectures and stochastic optimisers.


\vspace{-0.1cm}
\section{Related Work}
\vspace{-0.1cm}

%
The effects of example weighting and loss function are generally overlapped. 
Therefore, we need:
\\
\textbf{{Rethinking existing robustness theorems on loss functions}}.
In some prior work, when judging the robustness of a loss function, its  weighting scheme is not analysed. 
Instead, its robustness is judged according to its sensitivity to large errors \citep{huber1981robust,hastie2015statistical,rooyen2015learning,ghosh2017robust,charoenphakdee2019symmetric}. 
%
%
Especially in multi-class classification, \cite{ghosh2017robust} proposed theorems showing that a deep model is robust to label noise when the used loss function is symmetric and bounded. Accordingly, they claim that Mean Absolute Error (MAE), Mean Square Error (MSE), Categorical Cross Entropy (CCE) are decreasingly robust. However, in this work, we find CCE is very competitive with MAE, MSE and Generalised Cross Entropy (GCE) \citep{zhang2018generalized}. 
But CCE is neither bounded nor symmetric, thus being sensitive to large errors. This contrast indicates the theory may not be applicable for guiding real-world practice. 
\\
\textbf{{Rethinking proposed example weighting schemes}}. 
Many weighting schemes have been proposed for different purposes: 
(1) Easier examples are preferred and examples with lower error/loss are inferred to be easier~\citep{chang2017active,han2020sigua,yao2020searching}. For example, curriculum learning \citep{bengio2009curriculum} picks easier examples in early training. Self-paced learning \citep{kumar2010self,jiang2015self} increases the weights of more difficult examples gradually. 
(2) Harder examples are emphasised: Hard example mining is demonstrated to accelerate convergence and improve performance in some cases \citep{shrivastava2016training,gopal2016adaptive,loshchilov2016online}.
However, we remark a loss's derivative magnitude defines a weighting scheme. 
Then it becomes the interaction between a proposed weighting scheme and the one from a loss function works in those prior work. 
{Instead, in DM, there is only one weighting scheme.} 

\vspace{-0.1cm}
\section{Derivative Manipulation}
\label{sec:gradient_rescaling}
\vspace{-0.1cm}

Let a training set be $N$ training examples $\mathbf{X}=\{(\mathbf{x}_i, y_i)\}_{i=1}^N$, where $(\mathbf{x}_i, y_i)$ denotes $i$th sample with input $\mathbf{x}_i \in \mathbb{R}^D$ and label $y_i \in \{1,2, ..., C\}$. $C$ is the number of classes. Consider a deep neural network $z$ composed of an embedding network $f(\cdot): \mathbb{R}^D \rightarrow \mathbb{R}^K$ and a linear classifier $g(\cdot): \mathbb{R}^K \rightarrow \mathbb{R}^C$, i.e., $\mathbf{z}_i=z(\mathbf{x}_i)=g(f(\mathbf{x}_i)): \mathbb{R}^D \rightarrow \mathbb{R}^C$. Generally, a linear classifier is the last fully-connected layer which outputs logits $\mathbf{z} \in \mathbb{R}^C$. 
To predict the probabilities of $\mathbf{x}_i$ belonging to different classes, $\mathbf{z}$ is normalised by a softmax function: 
$p(j|\mathbf{x}_i)= 
{\exp (\mathbf{z}_{ij})}
/{ \sum\nolimits_{m=1}^{C} \exp (\mathbf{z}_{im}) },
\label{eq:softmax_normalisation}
$
where $p(j|\mathbf{x}_i)$ is the probability of $\mathbf{x}_i$ belonging to class $j$. 
For brevity, we define $p_i=p(y_i|\mathbf{x}_i)$.

We study CCE, MAE, MSE and GCE. They take only $p_i$ as input and have a minimum when $p_i$ is 1, which makes $p(j|\mathbf{x}_i), j \neq y_i$ be zeros automatically\footnote{In the standard cases of MAE and MSE, $p(j|\mathbf{x}_i), j \neq y_i$ are also taken as inputs and pushed towards zeros. However, this is unnecessary because purely maximising $p_i$ towards one naturally does that too.}.  
Their derivatives share one direction.   
\textit{They perform differently only due to the derivate magnitude, which theoretically proves it is a key.
}Therefore, we propose DM to systematically study derivative's magnitude with its direction kept the same as those losses.
We choose $L_1$ norm to measure the magnitude because its expression contains only $p_i$, being simpler to analyse. 
%
For clarity, we formally define the emphasis mode and variance: 
%
\\
\textbf{Definition 1} (\textit{Emphasis Mode} $\psi $). 
In DM, an example's weight $w_i$ is a function of $p_i$ merely: $w_i =  t(p_i)$, $t$ is a mapping from $p_i$ to $w_i$.  
Therefore, we define $ \psi = \argmax\nolimits_{p_i} w_i, \text{~~} \psi  \in [0, 1]$. 
%
%
\\
\textbf{Definition 2} (\textit{Emphasis Variance} $\sigma$). {It is the weight variance over all data points in a batch and is defined by the variance of their weights.
$\sigma = \mathrm{E}((w_i^\mathrm{} - \mathrm{E}(w_i^\mathrm{}))^2)$, where $\mathrm{E}(\cdot)$ denotes the expectation.
}


\vspace{-0.1cm} 
\subsection{The derivative direction and magnitude of common losses
}
\label{sec:preliminary}
\vspace{-0.1cm}

\textbf{CCE}.\footnote{The derivation details of all losses are given in the supplementary material.} 
For a given $(\mathbf{x}_i, y_i)$, CCE and its derivative with respect to $\mathbf{z}_{i}$ are:
	\begin{align}
	\begin{split}
	L_\mathrm{CCE}(\mathbf{x}_i, y_i)= -  \log p(y_i|\mathbf{x}_i) 
	\Rightarrow 
	\frac{\partial L_\mathrm{CCE}}{ \partial \mathbf{z}_{ij}} = 
	\begin{cases} 
	p(y_i|\mathbf{x}_i)-1
	\text{, } & j = y_i  \\
	p(j|\mathbf{x}_i)  
	\text{, } & j \neq y_i
	\end{cases}.
	\label{eq:CCE_loss_gradient}
	\end{split}
	\end{align}
With $L_1$ norm, we have $||\frac{\partial L_\mathrm{CCE}}{\partial \mathbf{z}_{i}}||_1= 2(1-p(y_i|\mathbf{x}_i))=2(1-p_i)$. 
The weight of $\mathbf{x}_i$ is $w_i^\mathrm{CCE}=||\frac{\partial L_\mathrm{CCE}}{\partial \mathbf{z}_{i}}||_1=2(1-p_i)$, meaning examples with smaller $p_i$ get higher weights.
\\
\textbf{MAE}. Similarly as above:
	\begin{align}
	\begin{split}
	L_\mathrm{MAE}(\mathbf{x}_i, y_i)
	= 1-p(y_i|\mathbf{x}_i)
	\Rightarrow
	\frac{\partial L_\mathrm{MAE}}{\partial \mathbf{z}_{ij}} = 
	\begin{cases} 
	p(y_i|\mathbf{x}_i)(p(y_i|\mathbf{x}_i)-1)
	\text{, } & j = y_i  \\
	p(y_i|\mathbf{x}_i)p(j|\mathbf{x}_i)  
	\text{, } & j \neq y_i
	\end{cases}.
	\label{eq:MAE_loss_gradient}
	\end{split}
	\end{align}
The weight of $\mathbf{x}_i$ is
$w_i^\mathrm{MAE}=||\frac{\partial L_\mathrm{MAE}}{\partial \mathbf{z}_{i}}||_1= 2p(y_i|\mathbf{x}_i)(1-p(y_i|\mathbf{x}_i))=2p_i(1-p_i)$.
\\
\textbf{MSE}. Similarly as above:
	\begin{align}
	\fontsize{9.5pt}{9.5pt}\selectfont
	\begin{split}
	L_\mathrm{MSE}(\mathbf{x}_i, y_i)
	= {(1-p(y_i|\mathbf{x}_i))}^2 
	\Rightarrow
	\frac{\partial L_\mathrm{MSE}}{\partial \mathbf{z}_{ij}} = 
	\begin{cases} 
	-2p(y_i|\mathbf{x}_i)(p(y_i|\mathbf{x}_i)-1)^2
	\text{,} & j = y_i  \\
	-2p(y_i|\mathbf{x}_i)(p(y_i|\mathbf{x}_i)-1)p(j|\mathbf{x}_i)  
	\text{,} & j \neq y_i
	\end{cases}.
	\label{eq:MSE_loss_gradient}
	\end{split}
	\end{align}
The weight of $\mathbf{x}_i$ is 
$w_i^\mathrm{MSE}=||\frac{\partial L_\mathrm{MSE}}{\partial \mathbf{z}_{i}}||_1= 4p(y_i|\mathbf{x}_i)(1-p(y_i|\mathbf{x}_i))^2=4p_i(1-p_i)^2$.
\\
\textbf{GCE}. Similarly as above:
	\begin{align}
	\begin{split}
	L_\mathrm{GCE}(\mathbf{x}_i, y_i)
	= \frac{1-p(y_i|\mathbf{x}_i)^q}{q}
	\Rightarrow
	\frac{\partial L_\mathrm{GCE}}{\partial \mathbf{z}_{ij}} = 
	\begin{cases} 
	p(y_i|\mathbf{x}_i)^{q} (p(y_i|\mathbf{x}_i)-1)
	\text{, } & j = y_i  \\
	p(y_i|\mathbf{x}_i)^{q}p(j|\mathbf{x}_i)  
	\text{, } & j \neq y_i
	\end{cases},
	\label{eq:GCE_loss_gradient}
	\end{split}
	\end{align}
where $q \in [0,1]$ is a hyperparameter. 
The weight of $\mathbf{x}_i$ is 
$w_i^\mathrm{GCE}=||\frac{\partial L_\mathrm{GCE}}{ \partial\mathbf{z}_{i}}||_1= 2p(y_i|\mathbf{x}_i)^q(1-p(y_i|\mathbf{x}_i))=2p_i^q(1-p_i)$.\\
\textbf{Derivative direction}.
We note $\frac{\partial L_\mathrm{CCE}}{\partial \mathbf{z}_{i}}$, $\frac{\partial L_\mathrm{MAE}}{\partial \mathbf{z}_{i}}$, $\frac{\partial L_\mathrm{MSE}}{\partial \mathbf{z}_{i}}$ and $\frac{\partial L_\mathrm{GCE}}{\partial \mathbf{z}_{i}}$ share the direction. Concretely: 
	\begin{align}
	\begin{split}
	\frac{\partial L_\mathrm{MAE}}{\partial \mathbf{z}_{i}} = p_i \times \frac{\partial L_\mathrm{CCE}}{\partial \mathbf{z}_{i}};
	\text{~~}
	\frac{\partial L_\mathrm{MSE}}{\partial \mathbf{z}_{i}} = 2p_i\times(1-p_i)\times \frac{\partial L_\mathrm{CCE}}{\partial \mathbf{z}_{i}};
	\text{~~}
	\frac{\partial L_\mathrm{GCE}}{\partial \mathbf{z}_{i}} = p_i^q \times \frac{\partial L_\mathrm{CCE}}{\partial \mathbf{z}_{i}}.
	\label{eq:derivative_share_direction}
	\end{split}
	\end{align}
	\textbf{Derivative magnitude}.
	We summarise the weighting schemes of all losses as follows: 
	\begin{align}
	\begin{split}
	&w_i^\mathrm{CCE}
	= 2(1-p_i) \Rightarrow \psi_\mathrm{CCE} = 0;
	\text{~~~~~~~~~~}
	w_i^\mathrm{MAE}
	= 2p_i(1-p_i) \Rightarrow \psi_\mathrm{MAE} = 0.5;
	\\
	&w_i^\mathrm{MSE}
	=4p_i(1-p_i)^2  \Rightarrow \psi_\mathrm{MSE} = \frac{1}{3};
	\text{~~~~}
	w_i^\mathrm{GCE}
	=2p_i^q(1-p_i)  \Rightarrow \psi_\mathrm{GCE} = \frac{q}{q+1}.
	\label{eq:derivative_focus_summary}
	\end{split}
	\end{align}
	\vspace{-0.1cm}
	\subsection{Defining an example weighting scheme by an emphasis density function}
	\label{sec:emphasis_density_function}
	\vspace{-0.1cm}
	\subsubsection{Example weighting via derivative manipulation}
	%
	%
	Common losses perform differently only due to the derivate magnitude as shown in the previous section.
	Therefore, we manipulate the derivative magnitude directly. 
	Concretely, given a weighting function $w_i^\mathrm{DM}$, we scale CCE's derivative by $w_i^\mathrm{DM}/(2(1-p_i))$:  
	\begin{align}
	\triangledown \mathbf{z}_{i} = w_i^\mathrm{DM}/(2(1-p_i)) \times \frac{\partial L_\mathrm{CCE}}{\partial \mathbf{z}_{i}}.
	\end{align} 
	Then the gradient magnitude of $\mathbf{z}_{i}$ is: $||\triangledown \mathbf{z}_{i}||_1 = ||w_i^\mathrm{DM}/(2(1-p_i)) \times \frac{\partial L_\mathrm{CCE}}{\partial \mathbf{z}_{i}}||_1=w_i^\mathrm{DM}$.
	\\
	Treating $p_i$ as a continuous variable, $w_i^\mathrm{DM}$ can be interpreted as an emphasis density function (EDF). 
	Correspondingly, the integral (area under the curve of $w_i^\mathrm{DM}$) between a range, e.g., $[\psi_\mathrm{DM}-\varDelta, \psi_\mathrm{DM}+\varDelta]$, denotes the accumulative weight of examples whose $p_i$ is in this range.  $2\varDelta$ is the length of this range, denoting the examples of interest. 
	%
	As $p_i \in [0,1]$, we normalise an EDF by its integral over $[0,1]$, termed a derivative normalisation (DN):
	\vspace{-0.15cm}
	\begin{align}
	\begin{split}
	h(w_i^\mathrm{DM})  =
	\frac{w_i^\mathrm{DM}}
	{ \int_{0}^{1} w_i^\mathrm{DM} d_{p_i}} \Rightarrow 
	 \int_{0}^{1} h(w_i^\mathrm{DM}) d_{p_i}=1 .
	\end{split}
	\label{eq:DM_normalised_expression}
	\end{align}
	%
	We name $w_i^\mathrm{DM}$ and $h$ weighting function and EDF, respectively. 
	In Eq.~(\ref{eq:DM_normalised_expression}), the DN operator is trivial. 
	Therefore, {for brevity, we  discuss the variants of $w_i^\mathrm{DM}$ instead of the normalised $h$.} 
	Next, we discuss how we express an example weighting function $w_i^\mathrm{DM}$ mathematically. 	 
	\subsubsection{Design of example weighting schemes}
	\label{sec:alternatives_emphasis_density_function}
	
	We derive $w_i^\mathrm{DM}$ according to the PDFs of probability distributions of the exponential family. 
	\\
	\textbf{Normal distribution variant}.  
	$\psi \geq 0$ denotes the emphasis mode while $\beta$ adjusts the variance: 
	\begin{align}
	\begin{split}
	v_\mathrm{ND}(w_i^\mathrm{DM}; \psi, \beta)  = \exp (-\beta p_i (p_i-2\psi) ).
	\end{split}
	\end{align} 
	\textbf{Exponential distribution variant}. Harder examples have larger (smaller) weights if $\beta>0$ ($\beta < 0$):   
	\begin{align}
	\begin{split}
	&v_\mathrm{ED}(w_i^\mathrm{DM}; \beta)  = \exp (\beta (1-p_i) ).
	\end{split}
	\end{align}
	\textbf{Beta distribution variant}. 
	It covers all weighting schemes of the common losses shown in Eq.~(\ref{eq:derivative_focus_summary}). We remark the difference of coefficients can be ignored since they are gone after DN. $\alpha, \eta \geq 0$.
	\begin{align}
	\fontsize{9.5pt}{9.5pt}\selectfont
	\begin{split}
	v_\mathrm{BD}(w_i^\mathrm{DM}; \alpha, \eta)  = 
	p_i^{\alpha-1} (1-p_i)^{\eta-1}
	 = 
	\begin{cases} 
	w_i^\mathrm{CCE}/2 = (1-p_i), & \alpha=1, \eta=2  \\
	w_i^\mathrm{MAE}/2 = p_i(1-p_i), & \alpha=2, \eta=2  \\
	w_i^\mathrm{MSE}/4	=p_i(1-p_i)^2, & \alpha=2, \eta=3  \\
	w_i^\mathrm{GCE}/2 = p_i^q(1-p_i), & \alpha=q+1, \eta=2  	
	\end{cases}
	\end{split}
	\end{align}
	Both emphasis mode and variance matter. 
	However, adjusting the variance is inconvenient in $v_\mathrm{BD}(w_i^\mathrm{DM}; \alpha, \eta)$.
	%
	Although $v_\mathrm{ND}(w_i; \psi, \beta)$ controls both of them, its mathematical generality to other weighting schemes is not good.  
	%
	%
	Therefore, we design the other variant of $w_i^\mathrm{DM}$ as follows:  
	\vspace{-0.06cm}
	\begin{align}
	\label{eq:Unified_Derivative}
	\begin{split}
	&w_i^\mathrm{DM}  =\exp(\beta  p_i^\lambda(1-p_i)), \lambda \geq 0 
	\Rightarrow
	\psi_\mathrm{DM} = \frac{\lambda}{\lambda+1} \in [0, 1),
	\end{split}
	\end{align}
	where $ \lambda \text{~and~} \beta$ are the parameters to control the emphasis mode and variance, respectively. 
	  
	\textbf{Design reasons}. 
	By varying $\lambda$ and $\beta$ in Eq~(\ref{eq:Unified_Derivative}), we can show that (1) 
	if $\lambda=0$, $w_i^\mathrm{DM}=\exp (\beta (1-p_i))$, it becomes the same as an exponential distribution variant $v_\mathrm{ED}(w_i^\mathrm{DM}; \beta) $; 
	(2) If $\lambda=1$, $w_i^\mathrm{DM}$  is a normal distribution variant; 
	(3) Eq~(\ref{eq:Unified_Derivative}) can be viewed as an exponential transformation of $v_\mathrm{BD}(w_i^\mathrm{DM}; \alpha, \eta)$, where  $\alpha=\lambda+1, \eta=2$ and scale it by $\beta$ followed by an exponential transformation;
	(4) $w_i^\mathrm{DM}$ is an extension of $w_i^\mathrm{GCE}$ by making $\lambda \geq 0$, linear scaling and exponential transformation. As a result, the emphasis variance can be easily adjusted based on tasks.  
	
	\vspace{-0.068cm}
	Finally, DM's loss expression is not an elementary function and can be represented as  $\int_{p_i}^{1} \frac{ w_i^\mathrm{DM} }{2 p_i (1-p_y)} dp_i $, which is unbounded and non-symmetric in multi-class cases. 

	\vspace{-0.0cm}
	\section{Experiments}
	\vspace{-0.1cm}
	To prove DM's value as a useful example weighting framework, we conduct diverse experiments.
	Apart from label noise, all real-world datasets are highly imbalanced, e.g., the number of videos per person ranges from 1 to 271 in MARS \citep{zheng2016mars}, while the number of images per class varies between 18,976 and 88,588 in Clothing 1M \citep{xiao2015learning}.
	In all experiments, we fix the random seed and do not apply any random computational accelerator for an entirely fair comparison.
	To search hyperparameters, we use a separate trusted validation set on Clothing 1M. We create one when it is unavailable on CIFAR-10/100 \citep{krizhevsky2009learning}: we train on 80\% training data which is corrupted in synthetic noisy cases and use 20\% clean training data for validation. After searching, we retrain a final model on the entire training data to fairly compare with some prior results.     

	\begin{figure}[!t]
		\vspace{-0.4cm}
		\centering
		\begin{subfigure}[h!]{0.34\textwidth}
			\centering
			\captionsetup{width=1.0\textwidth}
			\includegraphics[width=1\textwidth]{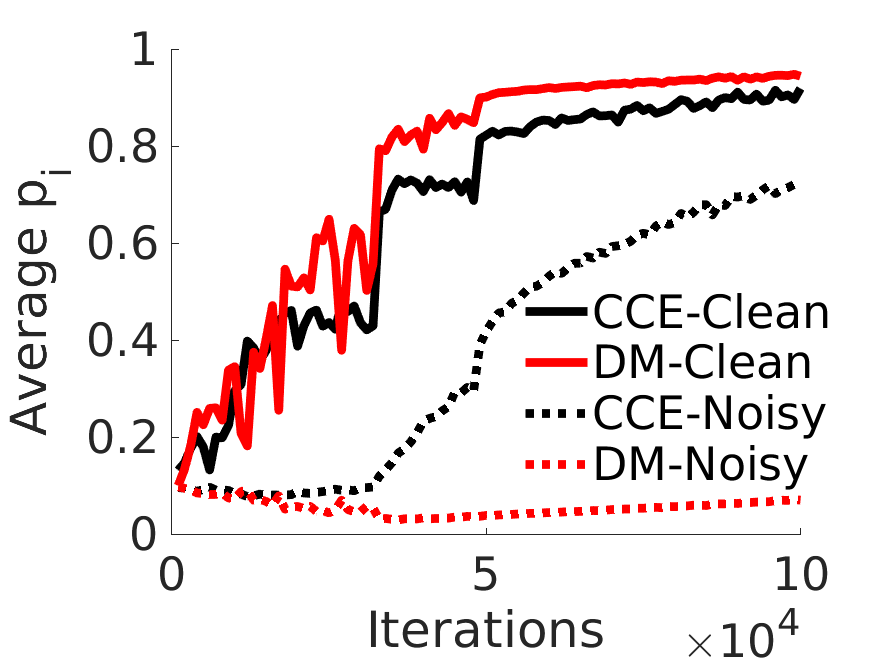}
			\caption{
				The average $p_i$ of different examples as training goes in CCE and DM.  
			}	
			\label{fig:dynamics_noisy}
		\end{subfigure}%
		\hfil
		\begin{subfigure}[h!]{0.34\textwidth}
			\vspace{-0.191cm}
			\centering
			\captionsetup{width=0.9\textwidth}
			\includegraphics[width=1\textwidth]{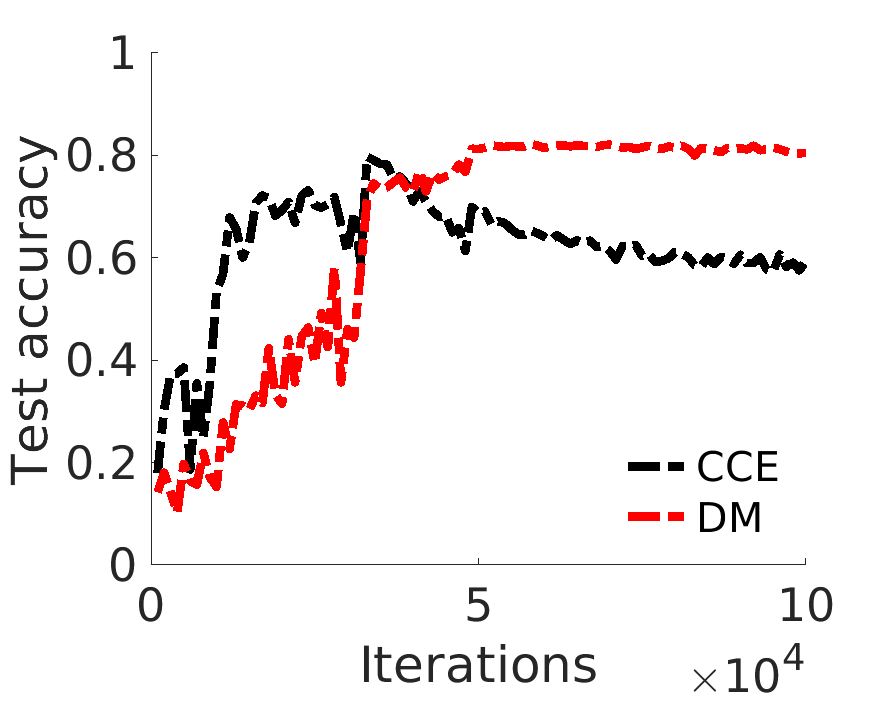}
			\caption{
				The test accuracy of CCE and DM as the iteration increases.     
			}
			\label{fig:dynamics_all}
		\end{subfigure}%
		\begin{subfigure}[h!]{0.32\textwidth}
			\vspace{-0.31cm}
			\centering
			\captionsetup{width=0.82\textwidth}
			\includegraphics[width=1\textwidth]{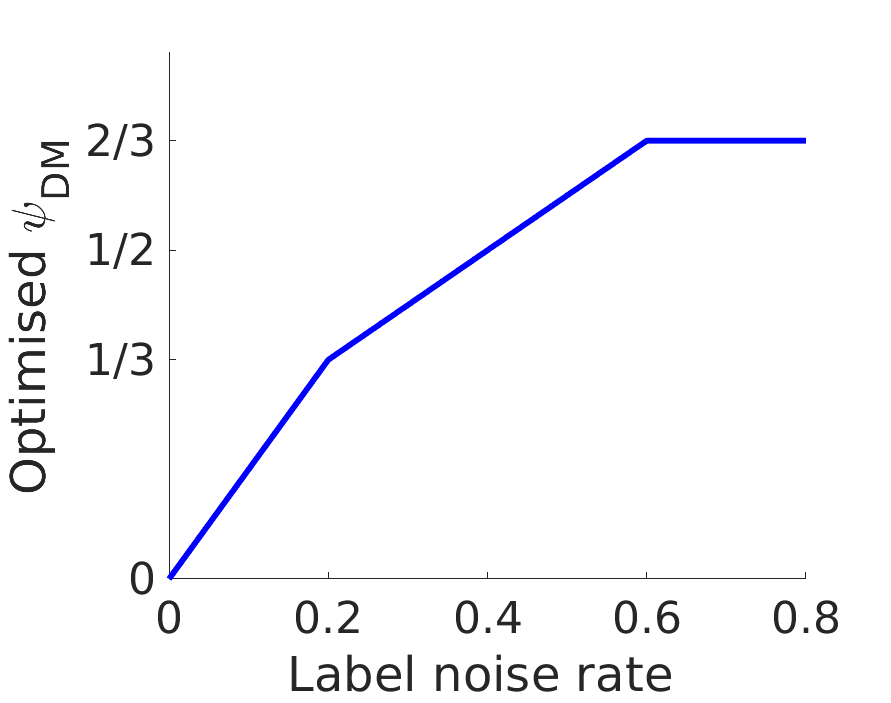}
			\caption{
				We optimise $\psi_\mathrm{DM}$ over four settings from $\{0,1/3, 1/2, 2/3\}$.     
			}
			\label{fig:EmphasisModeLabelNoise_v01}
		\end{subfigure}%
		\vspace{-5pt}
		\caption{
			We train ResNet-56 on CIFAR-10. 
			In (a) and (b), we observe noisy examples have \underline{much less $p_i$} than clean ones, thus being \underline{more difficult examples} in both CCE and DM. 
			In (c), for each label noise rate, we show the optimised $\psi_\mathrm{DM}$ from $\{0,1/3, 1/2, 2/3\}$, i.e., $\lambda \in \{0, 1/2, 1, 2\}$.  
		}
		\label{fig:dynamics_train_data}
		\vspace{-0.2cm}
	\end{figure}

	\vspace{-0.1cm}
	\subsection{Discussion on setting the emphasis mode when noisy labels exist}
	\label{sec:premise_justification_DM_helps}
	\vspace{-0.1cm}
	We optimise the emphasis mode according to an intuition: e.g., when there exists more noise, we use a relatively larger emphasis mode to emphasise on easier examples \citep{arpit2017closer,jiang2018mentornet}. 
	After an emphasis mode is set, we can search for the best emphasis variance on a validation set. Concretely, when training data is clean, we set $\lambda=0$ so that $w_i^\mathrm{DM}=\exp (\beta (1-p_i)) \Rightarrow \psi_\mathrm{DM}=0$. 
	When label noise exists, we increase $\lambda$ so that $\psi_\mathrm{DM}$ increases. For clarity, we summarise:
	
	\textbf{Premise 1}.  
	\textit{Difficult examples have smaller probabilities of being predicted to its annotated labels, i.e., smaller $p_i$.  	
	Abnormal examples, including noisy ones and outliers, belong to those difficult ones.} 
	
	This premise is justified in recent work:
	\citep{arpit2017closer} shows that DNNs do not fit real datasets by brute-force memorisation.
	Instead, DNNs learn simple shared patterns before memorising difficult abnormal data points. Consequently, abnormal samples have smaller $p_i$ than easier ones. 
	\\
	In addition, we present our empirical evidence in Figure~\ref{fig:dynamics_train_data}.  The learning dynamics prove this premise and DM is superior to CCE:  
	(a) The $p_i$ of clean examples increases while that of noisy ones has no noticeable rise in DM, which means DM hinders fitting of abnormal examples and preserves DNNs' ability to learn on clean data; 
	(b) DM has the best test accuracy. Furthermore, it is robust, i.e., early stopping is unnecessary.
	(c) As we increase the noise rate, the optimised emphasis mode also increases showing a positive correlation between them. A thorough ablation study on the emphasis mode and variance is reported in the Appendix~\ref{appendix_sec: ablation_study_onModeAndVariance}.	
	
	\textbf{Remark 1}. \textit{In the early training phase, $p_i$ is awfully random and non-informative, what does DM do?} 
	At this phase, DM randomly weights examples since $p_i$ is random. It is the same in all common losses (weighting schemes). 
	Although nothing makes sense at the beginning, including random initialisation, loss values and weighting schemes, DNNs gradually learn meaningful patterns. 

	\vspace{-0.1cm}
	\subsection{Robust image classification}
	\label{sec:robustness_cifar10_cifar100}
	\vspace{-0.1cm}
	
	\textbf{Datasets}.
	(1) \text{CIFAR-10/100} \citep{krizhevsky2009learning}, which contain 10 and 100 classes, respectively. 
	The image size is $32\times 32$. 
	In CIFAR-10, the training data contains 5k images per class while the test set includes 1k images per class. 
	CIFAR-100 has 500 images per class for training and 100 images per class for testing. 
	We generate synthetic label noise on them:  
	(a) {Symmetric noise.} With a probability of $r$, the original label of an image is changed to one of the other class labels uniformly following \citep{ma2018dimensionality,wang2019symmetric}. 
	$r$ denotes the noise rate. 
	We remark that some work randomly flips an image's label to one of all labels including the ground-truth \citep{tanaka2018joint,kim2019nlnl}. The actual noise rate becomes quite different when the number of classes is small, e.g., CIFAR-10. We do not compare with those results.   
	(b) {Asymmetric noise.} We generate asymmetric noise for CIFAR-100 following \citep{wang2019symmetric}. CIFAR-100 has 20 superclasses and every superclass has 5 subclasses. In each superclass, two subclasses are randomly selected  and their labels are flipped to each other with a probability of $r$. The overall noise rate is less than $r$.  
	(2) \text{Clothing 1M \citep{xiao2015learning}} is a real-world dataset and contains about 1 million images of 14 classes from shopping websites. 
	Its noise rate is about 38.46\% and noise distribution is agnostic. 

	\begin{table}[!t]
		\caption{
			Test accuracy (\%) on CIFAR-100.
			The best results on each block are bolded. 
			Italic row is the most basic baseline where examples have identical weights. 
		}
		\label{table:cifar100_SOTA_Symmetric}
		\centering
		\fontsize{8.6pt}{8.6pt}\selectfont
		\setlength{\tabcolsep}{6.0pt} 
		\vspace{-0.3cm}
		\begin{tabular}{ccccccccc}
			\toprule
			&\multirow{2}{*}{Method} & \multirow{2}{*}{\specialcell{Clean\\Labels}} & \multicolumn{3}{c}{Symmetric Noisy Labels} & \multicolumn{3}{c}{Asymmetric Noisy Labels} \\
			
			\cmidrule(l{2pt}r{2pt}){4-6}
			\cmidrule(l{2pt}r{2pt}){7-9}
			& & & $r$=0.2 & $r$=0.4 & \multicolumn{1}{c}{$r$=0.6}  & ~~~$r$=0.2 &~~ $r$=0.3 & $r$=0.4 \\
			
			\midrule

			\multirow{5}{*}{\specialcell{Results \\From \\ SL\\ \citep{wang2019symmetric}}}
			&LS & 63.7 & 58.8 & 50.1 & 24.7 &~~~ 63.0 & ~~62.3 & 61.6\\
			&Boot-hard & 63.3 & 57.9 & 48.2 & 12.3 & ~~~ 63.4 & ~~63.2 & 62.1\\
			&Forward & 64.0 & 59.8 & 53.1 & 24.7 &~~~ 64.1 & ~~64.0 & 60.9\\
			&D2L & 64.6 & 59.2 & 52.0 & 35.3 & ~~~62.4 & ~~63.2 & 61.4\\
			&SL & \textbf{66.8} & \textbf{60.0} &  \textbf{53.7} &  \textbf{41.5} 
			& ~~~\textbf{65.6} &  ~~\textbf{65.1} & \textbf{63.1} \\

			\midrule
			\multirow{13}{*}{\specialcell{Our \\Trained \\Results}}
			&CCE & \textbf{70.0} & {60.4} & {53.2} & {42.1} &~~~ \textbf{66.4} & ~~\textbf{64.7} & \textbf{60.3}\\
			&GCE & 63.6 & \textbf{62.4} & \textbf{58.6} & \textbf{50.6} &  ~~~62.8 & ~~62.2 & 58.7\\
			&MAE & {8.2} & {6.4} & {7.3} & {5.2} & ~~~7.3 & ~~6.3 & 7.3\\
			&MSE & 28.0 & 24.6 & 21.3 & 18.0 & ~~~ 24.5 & ~~24.3 & 23.0\\
			
			\cmidrule{2-9}
			&CCE-DN & \textbf{69.1} & {60.7} & {54.2} & {44.6} 
			& ~~~\textbf{65.9} & ~~\textbf{64.0} & \textbf{60.5}\\
			&GCE-DN & 65.8 & \textbf{62.5} & \textbf{58.3} & \textbf{48.4} 
			&  ~~~64.1 & ~~62.1 & 60.3\\
			&MAE-DN & {7.5} & {5.4} & {4.5} & {4.8} & ~~~5.8 & ~~3.5 & 3.9\\
			&MSE-DN & 25.8 & 28.4 & 27.0 & 26.5 & ~~~27.4 & ~~23.9 & 25.3\\
			
			\cmidrule{2-9}
			
			&\textit{DM($\beta=0$)} & \textit{67.2} & \textit{56.2} & \textit{50.9} & \textit{44.4} & ~~~\textit{64.4} & ~~\textit{62.5} & \textit{60.4} \\
			
			&DM($\lambda=0$) & \textbf{70.1} & {60.9} & {55.2} & {44.6} 
			& ~~~65.5 & ~~63.1 & 60.6\\
			
			&DM($\lambda=0.5$) & {69.3} & \textbf{65.7} & \textbf{61.0} & \textbf{52.9}
			& ~~~67.4 & ~~65.0 & 60.8 \\
			
			&DM($\lambda=1$) & {69.2} & {63.4} & {54.7} & {43.9} & ~~~\textbf{67.5} & ~~\textbf{65.8} &  \textbf{63.3}\\
			

			\bottomrule
		\end{tabular}
		\vspace{-0.3cm}
	\end{table}

	\vspace{-0.1cm}
	\subsubsection{Results on CIFAR-100}
	\label{sec:SOTA_CIFAR100}
	\vspace{-0.1cm}
	\textbf{Training details.}
	We follow the settings of \citep{wang2019symmetric} for a fair comparison.
	We use SGD with a momentum of 0.9 and a weight decay of $1e-4$. We train 30k iterations. The learning rate starts at 0.1, and is divided by 10 at 15k and 22k iterations. 
	Standard data augmentation is used: padding images with 4 pixels on every side, followed by a random crop of $32\times 32$ and horizontal flip. The batch size is 256.
	All methods use ResNet-44 \citep{he2016deep}. 
	%
	%
	\\
	\textbf{Competitors.} We briefly introduce the compared baselines:
	(1) Analysed losses (CCE, GCE, MAE and MSE) and their variants after DN (CCE-DN, GCE-DN, MAE-DN and MSE-DN); 
	(2) Bootstrapping trains a model with new labels generated by a convex combination of the original ones and their predictions. A convex combination can be either soft (Boot-soft) or hard (Boot-hard) \citep{reed2015training}; 
	(3) Forward (Backward) uses a noise-transition matrix to multiply the network's predictions (losses) for label correction \citep{patrini2017making};
	(4) D2L addresses noise-robustness by restricting the dimensionality expansion of learned subspaces during training;  
	(5) SL modifies CCE symmetrically with a reverse cross entropy;
	(6) LS denotes label smoothing \citep{hinton2015distilling}.
	%
	%
	Note that we do not compare with \citep{lee2019robust}. 
	First, its backbone is not ResNet-44 after checking with the authors. Second, their algorithm is orthogonal to ours because it targets at the inference stage and is a generative classifier on top of deep representations. 
	%
	%
	\\
	\textbf{Results.} 
	From Table~\ref{table:cifar100_SOTA_Symmetric}, 
	our observations are: 
	(1) When training data is clean, CCE (CCN-DN) is the best against other common losses.  
	Besides, DM($\lambda=0$) is the best compared with other variants. 
	We conclude by $\psi=0$, harder examples have higher weights, leading to better performance.
	(2) When label noise exists, we obtain better performance by increasing $\lambda$ so that $\psi$ is larger. 
	This demonstrates that we should focus on easier data points as label noise increases.

	\vspace{-0.1cm}
	\subsubsection{Results on CIFAR-10}
	\label{sec:SOTA_CIFAR10}
	\vspace{-0.1cm}
	\textbf{Training details.}
	We follow the same settings as MentorNet \citep{jiang2018mentornet} and train GoogLeNet V1 to compare fairly with its reported results. 
	Optimiser and data augmentation are the same as CIFAR-100. 
	\\
	\textbf{Competitors.}
	\textit{Self-paced \citep{kumar2010self}, Focal Loss \citep{lin2017focal}, and MentorNet are representatives of example weighting algorithms.}
	Forgetting \citep{arpit2017closer} searches the dropout parameter in the range of $(0.2, 0.9)$.  
	All methods use GoogLeNet V1 \citep{szegedy2015going}.
	\\
	\textbf{Results.} In Table~\ref{table:CIFAR10_more_baselines}, we observe: 
	(1) When looking at common losses, they perform differently in different cases. For example, CCE and MSE-DN are better when $r=0$ and $r=0.8$. GCE and GCE-DN are preferred when $r=0.2$ and $r=0.4$;
	(2) For DM, we first set $\lambda=0.5$, then optimise $\beta$ manually. We find it performs the best. $\beta=0$ denotes all samples have identical weights.

	\begin{table}
		\centering
		\parbox{.423\linewidth}{
			\captionsetup{width=.43\textwidth}
			\caption{
				Accuracy (\%) of DM and {other baselines} on CIFAR-10 under symmetric label noise. 
				The best results on each block are bolded.
				Number format of this table follows MentorNet. 
				`--' denotes result was not reported. 
			}
			\label{table:CIFAR10_more_baselines}
			\centering
			\fontsize{8.5pt}{8.5pt}\selectfont
			\setlength{\tabcolsep}{0.4pt} 
			\vspace{-0.24cm}
			\begin{tabular}{cccccc}
				\toprule
				&\makecell{Method}& \makecell{$r$=0} & \makecell{$r$=0.2} & \makecell{$r$=0.4} & \makecell{$r$=0.8} \\
				\midrule
				
				\multirow{7}{*}{\specialcell{Results From \\ MentorNet}}
				&\makecell{Forgetting} & -- & 0.76 & 0.71 & 0.44\\
				&\makecell{Self-paced} & -- & \textbf{0.80} & 0.74 & 0.33\\
				&\makecell{Focal Loss} & -- & 0.77 & 0.74 & 0.40\\
				&\makecell{Boot-soft} & -- & 0.78 & 0.73 & 0.39\\
				&\makecell{MentorNet PD} & -- & 0.79 & 0.74 & 0.44\\
				&\makecell{MentorNet DD} & -- & 0.79 & \textbf{0.76} & \textbf{0.46}\\
				\midrule
				
				\multirow{12}{*}{\specialcell{Our Trained \\ Results}}
				&\makecell{CCE} & \textbf{0.85} & 0.74 & 0.74 & \textbf{0.35}\\
				&\makecell{GCE} & 0.83 & \textbf{0.81} & \textbf{0.77} & 0.18\\
				&\makecell{MAE} & 0.57 & 0.50 & 0.45 & 0.19\\
				&\makecell{MSE} & 0.80 & 0.78 & 0.73 & 0.29\\
				\cmidrule{2-6}
				&\makecell{CCE-DN} & 0.84 & 0.75 & 0.76 & 0.18\\
				&\makecell{GCE-DN} & 0.77 & \textbf{0.82} & \textbf{0.79} & 0.19\\
				&\makecell{MAE-DN} & 0.10 & 0.10 & 0.10 & 0.19\\
				&\makecell{MSE-DN} & \textbf{0.85} & 0.51 & 0.76 & \textbf{0.53}\\
				\cmidrule{2-6}
				&\textit{{DM$(\beta=0$)}} & \textit{\textbf{0.86}} & \textit{0.66} & \textit{0.56} & \textit{0.18}\\
				&\makecell{DM$(\lambda=0.0$)} & \textbf{0.86} & 0.77 & 0.75 & 0.18\\
				&\makecell{DM$(\lambda=0.5$)} & \textbf{0.86} & \textbf{0.83} & \textbf{0.80} & \textbf{0.57}\\
				
				\bottomrule
			\end{tabular}
		}
		\hfill
		\parbox{.5\textwidth}{
			\centering
			\captionsetup{width=.5\textwidth}
			\caption{
				Experiments on sentiment classification of movie reviews.  
				Due to the space shortage, \textit{the results of variants after DN are in the brackets.} 
				We test on two adverse cases: label noise and sample imbalance. 
				P-N Ratio denotes the ratio of positive reviews to negative ones.  	
			}
			\label{table:IMDB}
			\fontsize{8.0pt}{8.0pt}\selectfont
			\setlength{\tabcolsep}{0.12pt} 
			\vspace{-0.3cm}
			\begin{tabular}{ccccccc}
				\toprule
				& &  \makecell{CCE(-DN)}~ & \makecell{GCE(-DN)}~ & \makecell{MAE(-DN)}~ & \makecell{MSE(-DN)}~ & DM\\
				\midrule
				
				\multirow{3}{*}{\specialcell{Label\\Noise\\ $r$}}
				&\makecell{~0.0~} & 88.9(88.5) & 88.9(87.7) & 88.9(74.5) & 88.9(88.2) & \textbf{89.1}\\ 
				&\makecell{~0.2~} & 87.7(87.2) & 88.6(85.5) & 88.6(72.8) & 88.5(87.0)&\textbf{88.7}\\
				&\makecell{~0.4~} & 75.5(75.0) & 83.6(75.2) & 84.9(80.2 ) & 83.7(75.3) & \textbf{86.4} \\ 
				\midrule
				\multirow{2}{*}{\specialcell{P-N\\ Ratio~}}
				&\makecell{10:1} & 78.9(77.5) & 77.0(59.1) & 75.4(0.5) & 77.6(78.5) & \textbf{80.6} \\ 
				&\makecell{50:1} & 63.4(61.4) & 0.5(0.5) & 0.5(0.5) & 58.6(64.4) & \textbf{65.0} \\ 
				\bottomrule
			\end{tabular}
			%
			%
			\vspace{0.20cm}
			\centering
			\captionsetup{width=.5\textwidth}
			\caption{
				Results of common stochastic optimisers.
				Adam \citep{kingma2015adam} is an adaptive gradient method.
				We report three settings of it. 	
			}
			\label{table:stochastic_optimisers}
			\fontsize{8.5pt}{8.5pt}\selectfont
			\setlength{\tabcolsep}{0.5pt} 
			\vspace{-0.3cm}
			\begin{tabular}{lccccc}
				\toprule
				&CCE & GCE & MAE & MSE & DM \\
				\midrule
				
				\makecell{SGD (lr: 0.01)} & 64.6 & 68.8 & 39.3 & 58.4 & \textbf{82.0}\\
				
				\makecell{SGD + Momentum  (lr: 0.01)} & 61.7 & 80.7 & 64.7 & 76.7 & \textbf{83.8} \\
				
				\makecell{Nesterov (lr: 0.01)} & 57.3 & 80.0 & 63.9 & 76.8 & \textbf{84.0}\\
				
				\midrule
				
				\makecell{Adam (lr: 0.01, delta: 0.1)} & 39.3 & 75.7 & 57.5 & 66.8 & \textbf{78.2}\\
				
				\makecell{Adam (lr: 0.005, delta: 0.1)}  & 44.3 & 72.6 & 60.8 & 67.9 & \textbf{80.8} \\
				
				\makecell{Adam (lr: 0.005, delta: 1)} & 52.0 & 67.7 & 37.3 & 58.5 & \textbf{79.2} \\
				
				\bottomrule
			\end{tabular}
			%
		}
		\vspace{-0.3cm}
	\end{table}

	\vspace{-0.1cm}
	\subsubsection{Results on Clothing 1M}
	\vspace{-0.1cm}
	
	\textbf{Training details.}
	We follow \citep{tanaka2018joint} to train ResNet-50 \citep{he2016deep}: 
	(1) We initialise it by a pretrained model on ImageNet \citep{russakovsky2015imagenet}; 
	(2) SGD with a momentum of $0.9$ and a weight decay of $2e{-5}$ is applied. The learning rate starts at $0.01$ and is divided by 10 at 10k and 15k iterations. We train 20k iterations;	(3) Data augmentation: first resize a raw input image to $256 \times 256$, and then crop it randomly at $224 \times 224$ followed by random horizontal flipping. 
	We set $\lambda=1,\beta=2$ for DM.
	\\
	\textbf{Competitors.} We compare with recent algorithms:
	(1) S-adaptation applies an auxiliary softmax layer to estimate a noise-transition matrix \citep{goldberger2017training}; 
	(2) Masking is a human-assisted approach that conveys human cognition to speculate the structure of a noise-transition matrix \citep{han2018masking};  
	(3) Joint Optim. \citep{tanaka2018joint} learns latent true labels and model's parameters iteratively. Two regularisation terms are added for label estimation and adjusted in practice;
	(4) MD-DYR-SH \citep{arazo2019unsupervised} combines dynamic mixup (MD), dynamic bootstrapping plus regularisation (DYR) from soft to hard (SH).   
	\\
	\textbf{Results.} In Table~\ref{table:Clothing1M_competitors}, under real-world agnostic noise, DM outperforms the state-of-the-art. 
	It is worth noting that the burden of noise-transition matrix estimation in Forward, S-adaptation, Masking and Joint Optim. is heavy, whereas DM is simple and effective.

	\begin{table}[!t]
		\caption{
			Accuracy (\%) on Clothing1M. 
			The leftmost block's results are from SL while the middle block's are from Masking.
			Results of corresponding variants after DN are in the brackets.  
		}
		\centering
		\setlength{\tabcolsep}{1.15pt} 
		\fontsize{8.1pt}{8.1pt}\selectfont
		\vspace{-0.3cm}
		\begin{tabular}{cccc|cc|ccccccc}
			\toprule
			
			 \multirow{2}{*}{\makecell{Boot-\\hard}} & \multirow{2}{*}{{D2L} } & 
			\multirow{2}{*}{{Forward}} &  
			\multirow{2}{*}{{SL~~}} & 
			\multirow{2}{*}{\makecell{~S-\\~~~adaptation}} &
			\multirow{2}{*}{ Masking~~} &
			\multirow{2}{*}{ \makecell{~Joint \\~Optim.}} &
			\multirow{2}{*}{ \makecell{~MD-~~\\DYR-~~\\SH~~}} &
			\multicolumn{5}{c}{Our Trained Results}  
			\\
			\cmidrule{9-13}
			
			&  &   &  &  &  &  & & CCE(-DN) & GCE(-DN) & MAE(-DN) & MSE(-DN) & DM\\
			\midrule
			68.9 & 69.5  & 69.8 & 71.0~~  
			& 70.3 & 71.1~~ & 72.2 & 71.0 & 71.7(72.5) &  {72.4(64.5)} & 39.7(16.4) & 71.7(69.9) & \textbf{73.3} \\
			\bottomrule
		\end{tabular}
		\label{table:Clothing1M_competitors}
		\vspace{-0.2cm}
	\end{table}

	\vspace{-0.1cm}
	\subsection{Robust video retrieval}
	\vspace{-0.1cm}
	
	\noindent
	\textbf{Dataset and evaluation settings.} MARS \citep{zheng2016mars} contains 20,715 videos of 1,261 persons. 
	There are 1,067,516 frames in total. 
	Because person videos are collected by tracking and detection algorithms, 
	abnormal examples exist as shown in Figure~\ref{fig:abnormal_examples} in the supplementary material: \textit{Some frames contain only background or an out-of-distribution person.} 
	\textit{Exact noise type and rate are unknown.}
	We use 8,298 videos of 625 persons for training 
	and 12,180 videos of the other 636 persons for testing. 
	We report the cumulated matching characteristics (CMC) and mean average precision (mAP) results \citep{zheng2016mars}.
	\\
	\noindent
	\textbf{Implementation details.
	}  Following \citep{liu2017qan,wang2019deep}, we train GoogleNet V2 \citep{ioffe2015batch} and process a video as an image set, which means we use only appearance information without exploiting latent temporal information. 
	A video's representation is simply the average fusion of its frames' representations.
	%
	The learning rate starts from 0.01 and is divided by 2 every 10k iterations. We stop training at 50k iterations. 
	We apply an SGD optimiser with a weight decay of $5e-4$ and a momentum of 0.9. The batch size is 180. 
	Data augmentation is the same as Clothing 1M.
	%
	At testing, we $L_2$ normalise videos' features and calculate the cosine similarity.  
	%
	\\
	\noindent
	\textbf{Results.} 
	The results are displayed in Table~\ref{table:MARS_ReID}.
	%
	Although DRSA \citep{li2018diversity} and CAE \citep{chen2018video} exploit extra temporal information by incorporating attention mechanisms, DM is superior to them in terms of both effectiveness and simplicity. OSM+CAA \citep{wang2019deep} is the only competitive method. However, OSM+CAA combines CCE and weighted contrastive loss to address anomalies, thus being more complex. 
	We highlight that one query may have multiple matching instances in the MARS benchmark so that mAP is a more reliable and accurate performance assessment. DM is the best in terms of mAP.   
	
	\begin{table}[!t]
		\vspace{0.0cm}
		\caption{
			Video retrieval results on MARS dataset. All other methods use GoogLeNet V2 except that DRSA and CAE use more complex ResNet-50. 
		}
		\centering
		\fontsize{8.2pt}{8.2pt}\selectfont
		\setlength{\tabcolsep}{6.0pt} 
		\vspace{-0.3cm}
		\begin{tabular}{lcccccccc}
			\toprule
			\multirow{2}{*}{Metric}  & \multirow{2}{*}{DRSA} & \multirow{2}{*}{CAE} & \multirow{2}{*}{\specialcell{~OSM+CAA~~~}}
			& \multicolumn{5}{c}{Our Trained Results}
			\\
			\cmidrule{5-9}
			& & & & CCE & GCE & MAE & MSE & DM \\
			\midrule
			mAP (\%) & 65.8 & 67.5 & 72.4& 58.1 & 31.6  & 12.0 & 19.6 & \textbf{72.8}\\
			CMC-1 (\%) & 82.3 & 82.4 & \textbf{84.7} & 73.8 & 51.5 & 26.0 & 39.3 & {84.3}\\
			\bottomrule
		\end{tabular}
		\label{table:MARS_ReID}
		\vspace{-0.2cm}
	\end{table}
	

	\vspace{-0.1cm}
	\subsection{Sentiment analysis of movie reviews}
	\vspace{-0.1cm}
	We report results on the IMDB dataset of movie reviews \citep{maas2011learning, mesnil2014ensemble}. 
	We use Paragraph Vector, PV-DBOW, as a document descriptor \citep{le2014distributed}. We train a neural network with one 8-neuron hidden layer and display the results in Table~\ref{table:IMDB}. 
	Due to space limitation, other details are reported in the Appendix~\ref{appendix_sec: movie_review}. 
%

	\vspace{-0.1cm}
	\subsection{Further analysis}
	\vspace{-0.1cm}
	We also experimented with Adam optimiser as shown in Table~\ref{table:stochastic_optimisers}. 
	To explore different networks simultaneously, we train ResNet-56 \citep{he2016deep} instead of GoogLeNet V1 on CIFAR-10 with 40\% symmetric label noise. 
	We observe that DM's results are consistently the best. Additionally, experiments about comparison with standard regularisers, performance on a small-scale dataset and detailed training analysis under label noise are presented in the Appendixes~\ref{appendix_sec:standard_regularisers}, \ref{appendix_sec:small_scale_fine_grained} and \ref{appendix_sec: ablation_study_onModeAndVariance}, respectively.  
	
	\vspace{-0.1cm}
	\section{Conclusion and Future Work}
	\vspace{-0.1cm}

	In this work, we propose derivative manipulation for example weighting. DM directly works on gradients bypassing a loss function. 
	As a consequence, it creates great flexibility in designing various example weighting schemes. 
	Extensive experiments on both vision and language tasks empirically show that DM outperforms existing methods despite its simplicity. 

	For the future work, firstly, we will pay attention to how to set the emphasis mode and variance adaptively on different datasets and progressively along with training time. Secondly, some other losses have different derivative directions, e.g., when target modification is applied \citep{wang2020proselflc}. 
	We acknowledge that this is also a valuable area for future research, e.g., how to manipulate the derivative direction.

\bibliography{ICML2020_DM}
\bibliographystyle{iclr2021_conference}

\newpage

\appendix

\begin{center}
	\Large
	\textbf{Supplementary Material of Derivative Manipulation}
\end{center}

\maketitle


\section{Display of Semantically Abnormal Training Examples}

\begin{figure*}[!h]
	\centering
	\includegraphics[width=0.55\linewidth]{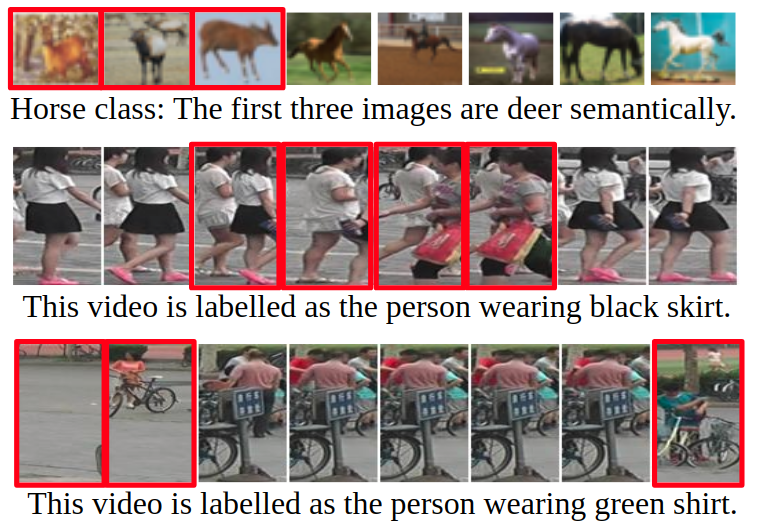}
	\caption{
		Diverse semantically abnormal training examples highlighted by red boxes. 
		The 1st row shows synthetic abnormal examples from corrupted CIFAR-10
		~\citep{krizhevsky2009learning}.
		The 2nd and 3rd rows present realistic abnormal examples from video person re-identification benchmark MARS \citep{zheng2016mars}.    
		\\
		\textit{\textbf{Out-of-distribution anomalies}}: 1) The first image in the 3rd row contains only background  and no semantic information at all. 2) The 2nd first image or the last one in the 3rd row may contain a person that does not belong to any person in the training set. 
		\\
		\textit{\textbf{In-distribution anomalies}}: 
		1) Some images of deer class are wrongly annotated to horse class. 
		2) We cannot decide the object of interest without any prior when an image contains more than one object, e.g., some images contain two persons in the 2nd row.		
	}
	\label{fig:abnormal_examples}
\end{figure*}


\section{Derivation Details of Softmax, CCE, MAE and GCE}
\label{sec:derivation_details}

\subsection{Derivation of Softmax Normalisation}
We rewrite $p(y_i|\mathbf{x}_i)$ as follows:
\begin{equation}
	\label{eq:predict_probability}
	\begin{aligned}
		%
		p(y_i|\mathbf{x}_i)^{-1}
		&= 1+\sum_{j\neq y_i} \exp(\mathbf{z}_{ij}-\mathbf{z}_{iy_i})
		. 
	\end{aligned}
\end{equation}


\noindent
For left and right sides of Eq.~(\ref{eq:predict_probability}), we calculate their derivatives w.r.t. $\mathbf{z}_{ij}$ simultaneously.

If $j = y_i$, 
\begin{equation}
	\label{eq:both_prob_derivation_final_y_i}
	\begin{aligned}
		&\frac{-1}{p(y_i|\mathbf{x}_i)^2} 
		\frac{\partial p(y_i|\mathbf{x}_i)}{\mathbf{z}_{iy_i}} 
		= 
		-\sum_{j\neq y_i} \exp(\mathbf{z}_{ij}-\mathbf{z}_{iy_i})
		\\
		&=>
		\frac{\partial p(y_i|\mathbf{x}_i)}{\mathbf{z}_{iy_i}} 
		= 
		p(y_i|\mathbf{x}_i) 
		(1-p(y_i|\mathbf{x}_i))
		. 
	\end{aligned}
\end{equation}


\noindent
If $j \neq y_i$,

\begin{equation}
	\label{eq:both_prob_derivation_final_j}
	\begin{aligned}
		&\frac{-1}{p(y_i|\mathbf{x}_i)^2} 
		\frac{\partial p(y_i|\mathbf{x}_i)}{\mathbf{z}_{ij}} = 
		\exp(\mathbf{z}_{ij}-\mathbf{z}_{iy_i})\\
		&=>
		\frac{\partial p(y_i|\mathbf{x}_i)}{\mathbf{z}_{ij}} 
		= 
		-p(y_i|\mathbf{x}_i) 
		p(j|\mathbf{x}_i)
		. 
	\end{aligned}
\end{equation}

\noindent
\textit{In summary}, the derivation of softmax layer is:
\begin{equation}
	\label{eq:both_prob_derivation_final}
	\begin{aligned}
		\frac{\partial p(y_i|\mathbf{x}_i)}{\partial \mathbf{z}_{ij}} 
		&=
		\begin{cases} 
			p(y_i|\mathbf{x}_i) 
			(1-p(y_i|\mathbf{x}_i))
			\text{, } &j = y_i  \\
			-p(y_i|\mathbf{x}_i) 
			p(j|\mathbf{x}_i)  
			\text{, } &j \neq y_i
		\end{cases}
	\end{aligned}
\end{equation}

\subsection{Derivation of CCE}
According to Eq.~(\ref{eq:CCE_loss_gradient}), 
we have
\begin{equation}
	\label{eq:loss_CCE_x_i}
	\begin{aligned}
		L_{\mathrm{CCE}} (\mathbf{x}_i;f_\theta,\mathbf{W}) 
		&= 	- \log p(y_i|\mathbf{x}_i)
		.
	\end{aligned}
\end{equation}
Therefore, we obtain (the parameters are omitted for brevity),   
\begin{equation}
	\label{eq:derivation_CCE_x_i}
	\begin{aligned}
		\frac{\partial L_{\mathrm{CCE}}}{\partial p(j|\mathbf{x}_i)} 
		&=
		\begin{cases} 
			-p(y_i|\mathbf{x}_i)^{-1} 
			\text{, } &j = y_i  \\
			0       
			\text{, } &j \neq y_i
		\end{cases}
		.
	\end{aligned}
\end{equation}


\subsection{Derivation of MAE}
According to Eq.~(\ref{eq:MAE_loss_gradient}), 
we have
\begin{equation}
	\label{eq:loss_MAE_x_i}
	\begin{aligned}
		L_{\mathrm{MAE}} (\mathbf{x}_i;f_\theta,\mathbf{W}) 
		&= 	1- (p(y_i|\mathbf{x}_i)
		.
	\end{aligned}
\end{equation}
Therefore, we obtain 
\begin{equation}
	\label{eq:derivation_MAE_x_i}
	\begin{aligned}
		\frac{\partial L_{\mathrm{MAE}}}{\partial p(j|\mathbf{x}_i)} 
		&=
		\begin{cases} 
			-1  \text{, } &j = y_i  \\
			0        \text{, } &j \neq y_i
		\end{cases}
		.
	\end{aligned}
\end{equation}

\subsection{Derivation of GCE}
According to Eq.~(\ref{eq:GCE_loss_gradient}), 
we have
\begin{equation}
	\label{eq:loss_GCE_x_i}
	\begin{aligned}
		L_{\mathrm{GCE}} (\mathbf{x}_i;f_\theta,\mathbf{W}) 
		&= 	\frac{1-p(y_i|\mathbf{x}_i)^q}{q}
		.
	\end{aligned}
\end{equation}
Therefore, we obtain 
\begin{equation}
	\label{eq:derivation_GCE_x_i}
	\begin{aligned}
		\frac{\partial L_{\mathrm{GCE}}}{\partial p(j|\mathbf{x}_i)} 
		&=
		\begin{cases} 
			-p(y_i|\mathbf{x}_i)^{q-1}  \text{, } &j = y_i  \\
			0        \text{, } &j \neq y_i
		\end{cases}
		.
	\end{aligned}
\end{equation}

\subsection{Derivatives w.r.t. Logits $\mathbf{z}_{i}$}


\subsubsection{{$\partial L_{\mathrm{CCE}}/ \partial \mathbf{z}_{i}$}}
The calculation is based on Eq.~(\ref{eq:derivation_CCE_x_i}) and Eq.~(\ref{eq:both_prob_derivation_final}).

\noindent
If $j = y_i$, we have:
\begin{equation}
	\begin{aligned}
		\frac{\partial L_{\mathrm{CCE}}}{\partial \mathbf{z}_{iy_i}} 
		&= \sum_{j=1}^{C}  \frac{\partial L_{\mathrm{CCE}}}{\partial p(j|\mathbf{x}_i)}  \frac{\partial p(y_i|\mathbf{x}_i)}{\mathbf{z}_{ij}} \\
		&=p(y_i|\mathbf{x}_i)-1
		.
	\end{aligned}
\end{equation}

\noindent
If $j \neq y_i$, it becomes:
\begin{equation}
	\begin{aligned}
		\frac{\partial L_{\mathrm{CCE}}}{\partial \mathbf{z}_{ij}} 
		&= \sum_{j=1}^{C}  \frac{\partial L_{\mathrm{CCE}}}{\partial p(j|\mathbf{x}_i)}  \frac{\partial p(y_i|\mathbf{x}_i)}{\mathbf{z}_{ij}} \\
		&=p(j|\mathbf{x}_i)
		.
	\end{aligned}
\end{equation}

\noindent
\textit{In summary}, $\partial L_{\mathrm{CCE}}/ \partial \mathbf{z}_{i}$ can be represented as:
\begin{equation}
	\label{eq:summary_CCE_z}
	\begin{aligned}
		\frac{\partial L_{\mathrm{CCE}}}{\partial \mathbf{z}_{ij}} 
		&=
		\begin{cases} 
			p(y_i|\mathbf{x}_i)-1  \text{, } &j = y_i  \\
			p(j|\mathbf{x}_i)        \text{, } &j \neq y_i
		\end{cases}
		.
	\end{aligned}
\end{equation}

\subsubsection{$\partial L_{\mathrm{MAE}}/ \partial \mathbf{z}_{i}$} 
The calculation is analogous with that of  $\partial L_{\mathrm{CCE}}/ \partial \mathbf{z}_{i}$.

According to Eq.~(\ref{eq:derivation_MAE_x_i}) and Eq.~(\ref{eq:both_prob_derivation_final}), 
if $j = y_i$:
\begin{equation}
	\begin{aligned}
		\frac{\partial L_{\mathrm{MAE}}}{\partial \mathbf{z}_{iy_i}} &= \sum_{j=1}^{C}  \frac{\partial L_{\mathrm{MAE}}}{\partial p(j|\mathbf{x}_i)} 
		\frac{\partial p(y_i|\mathbf{x}_i)}{\mathbf{z}_{ij}} \\
		&= - p(y_i|\mathbf{x}_i) 
		(1-p(y_i|\mathbf{x}_i))
		.
	\end{aligned}
\end{equation}
otherwise ($j \neq y_i$):
\begin{equation}
	\begin{aligned}
		\frac{\partial L_{\mathrm{MAE}}}{\partial \mathbf{z}_{ij}} &= \sum_{j=1}^{C}  \frac{\partial L_{\mathrm{MAE}}}{\partial p(j|\mathbf{x}_i)} 
		\frac{\partial p(y_i|\mathbf{x}_i)}{\mathbf{z}_{ij}} \\
		&=  p(y_i|\mathbf{x}_i) 
		p(j|\mathbf{x}_i)
		.
	\end{aligned}
\end{equation}

\noindent
\textit{In summary}, $\partial L_{\mathrm{MAE}}/ \partial \mathbf{z}_{i}$ is:
\begin{equation}
	\label{eq:summary_MAE_z}
	\begin{aligned}
		\frac{\partial L_{\mathrm{MAE}}}{\partial \mathbf{z}_{ij}} 
		&=
		\begin{cases} 
			 p(y_i|\mathbf{x}_i)
			(p(y_i|\mathbf{x}_i)-1)  \text{, } &j = y_i  \\
			 p(y_i|\mathbf{x}_i)
			p(j|\mathbf{x}_i)        \text{, } &j \neq y_i
		\end{cases}
		.
	\end{aligned}
\end{equation}

\subsubsection{{$\partial L_{\mathrm{GCE}}/ \partial \mathbf{z}_{i}$}}
The calculation is based on Eq.~(\ref{eq:derivation_GCE_x_i}) and Eq.~(\ref{eq:both_prob_derivation_final}).

\noindent
If $j = y_i$, we have:
\begin{equation}
	\begin{aligned}
		\frac{\partial L_{\mathrm{GCE}}}{\partial \mathbf{z}_{iy_i}} 
		&= \sum_{j=1}^{C}  \frac{\partial L_{\mathrm{GCE}}}{\partial p(j|\mathbf{x}_i)}  \frac{\partial p(y_i|\mathbf{x}_i)}{\mathbf{z}_{ij}} \\
		&=p(y_i|\mathbf{x}_i)^{q}  (p(y_i|\mathbf{x}_i)-1)
		.
	\end{aligned}
\end{equation}

\noindent
If $j \neq y_i$, it becomes:
\begin{equation}
	\begin{aligned}
		\frac{\partial L_{\mathrm{GCE}}}{\partial \mathbf{z}_{ij}} 
		&= \sum_{j=1}^{C}  \frac{\partial L_{\mathrm{GCE}}}{\partial p(j|\mathbf{x}_i)}  \frac{\partial p(y_i|\mathbf{x}_i)}{\mathbf{z}_{ij}} \\
		&= p(y_i|\mathbf{x}_i)^{q}  p(j|\mathbf{x}_i)
		.
	\end{aligned}
\end{equation}

\noindent
\textit{In summary}, $\partial L_{\mathrm{GCE}}/ \partial \mathbf{z}_{i}$ can be represented as:
\begin{equation}
	\label{eq:summary_CCE_z}
	\begin{aligned}
		\frac{\partial L_{\mathrm{GCE}}}{\partial \mathbf{z}_{ij}} 
		&=
		\begin{cases} 
			p(y_i|\mathbf{x}_i)^{q}  (p(y_i|\mathbf{x}_i)-1)  \text{, } &j = y_i  \\
			p(y_i|\mathbf{x}_i)^{q} p(j|\mathbf{x}_i)        \text{, } &j \neq y_i
		\end{cases}
		.
	\end{aligned}
\end{equation}

\section{Beating Standard Regularisers Under Label Noise}
\label{appendix_sec:standard_regularisers}

In Table~\ref{table:CIFAR100_regulariser_competitors}, we compare our proposed DM  with other standard ones, i.e., L2 weight decay and Dropout \citep{srivastava2014dropout}.  
We set the dropout rate to 0.2 and L2 weight decay rate to $10^{-4}$.
For DM, 
we fix $\beta=8, \lambda=0.5$.
%
%
DM is better than those standard regularisers and their combinations significantly.  		
DM works best when it is together with L2 weight decay. 
%

\begin{table*}[!h]
	\caption{
		Results of DM and other standard regularisers on CIFAR-100. We set $r=40\%$, i.e., the label noise is severe but not belongs to the majority. We train ResNet-44. We report the average test accuracy and standard deviation (\%) over 5 trials.  
		Baseline is CCE without regularisation.  
	}
	\centering
	\setlength{\tabcolsep}{3.80pt} 
	\vspace{-6pt}
	\fontsize{9pt}{9pt}\selectfont
	\begin{tabular}{lccccccc}
		\toprule
		Baseline & L2  & Dropout & Dropout+L2 & DM & DM+L2 & DM+Dropout & DM+L2+Dropout\\
		\midrule
		44.7$\pm$0.1
		& 51.5$\pm$0.4
		& 46.7$\pm$0.5
		& 52.8$\pm$0.4
		& 55.7$\pm$0.3
		& \textbf{59.3}$\pm$0.2
		& 54.3$\pm$0.4
		& 58.3$\pm$0.3
		\\
		\bottomrule
	\end{tabular}
	\label{table:CIFAR100_regulariser_competitors}
\end{table*}

\section{Small-scale Fine-grained Visual Categorisation of Vehicles}
\label{appendix_sec:small_scale_fine_grained}

How does DM perform on small datasets, for example, the number of data points is no more than 5,000? 
We have tested DM on CIFAR-10 and CIFAR-100 in the main paper. However, both of them contain a training set of 50,000 images. 

For this question, we answer it from different perspectives as follows: 

\textit{1. The problem of label noise on CIFAR-10 and CIFAR-100 in Section~\ref{sec:robustness_cifar10_cifar100} is of similar scale.} 
\begin{itemize}
	\item In Table~\ref{table:CIFAR10_more_baselines}, when noise rate is 80\% on CIFAR-10, the number of clean training examples is around $50,000 \times 20\%=5,000\times 2$. Therefore, this clean set is only two times as large as 5,000. Beyond, the learning process may be interrupted by other noisy data points. 
	\item In Table~\ref{table:cifar100_SOTA_Symmetric}, when noise rate is 60\% on CIFAR-100, the number of clean training data points is about $50,000 \times 40\% = 5,000 \times 4$, i.e., four times as large as 5,000.    
\end{itemize}

\textit{2. We compare DM with other standard regularisers on a small-scale fine-grained visual categorisation problem in Table~\ref{table:Vehicles10_regulariser_competitors}.} 

\textbf{Vehicles-10 Dataset.} In CIFAR-100 \citep{krizhevsky2009learning}, there are 20 coarse classes, including vehicles 1 and 2. Vehicles 1 contains 5 fine classes: bicycle, bus, motorcycle, pickup truck, and train. Vehicles 2 includes another 5 fine classes: lawn-mower, rocket, streetcar, tank, and tractor. We build a small-scale vehicle classification dataset composed of these 10 vehicles from CIFAR-100. Specifically, the training set contains 500 images per vehicle class while the testing set has 100 images per class. Therefore, the number of training data points is 5,000 in total.

\begin{table*}[!h]
	\caption{
		The test accuracy (\%) of DM and other standard regularisers on Vehicles-10. We train ResNet-44.  
		Baseline denotes CCE without regularisation.  
		We test two cases: symmetric label noise rate is $r=40\%$, and clean data $r=0$. 
	}
	\centering
	\setlength{\tabcolsep}{6pt} 
	\vspace{-6pt}
	\fontsize{9pt}{9pt}\selectfont
	\begin{tabular}{ccccccccc}
		\toprule
		$r$ &Baseline & L2  & Dropout & Dropout+L2 & DM & DM+L2 & DM+Dropout & DM+L2+Dropout\\
		\midrule
		0& 75.4
		& 76.4
		& 77.9
		& 78.7
		& 83.8
		& 84.4
		& 84.5
		& \textbf{84.7}
		\\
		40\%& 42.3
		& 44.8
		& 41.6
		& 47.4
		& 45.8
		& 55.7
		& 48.8
		& \textbf{58.1}
		\\
		\bottomrule
	\end{tabular}
	\label{table:Vehicles10_regulariser_competitors}
\end{table*}

\section{Experimental Details of Robust Sentiment Analysis of Movie Reviews}
\label{appendix_sec: movie_review}

We report results on the IMDB dataset of movie reviews \citep{maas2011learning} following \citep{mesnil2014ensemble}. 
Paragraph Vector (PV-DBOW) is used as a document descriptor \citep{le2014distributed}. We train a neural network with one 8-neuron hidden layer and display the results in Table~\ref{table:IMDB}. 
We generate symmetric label noise. 


IMDB contains 25,000 positive movie reviews and 25,000 negative ones. We follow \citep{maas2011learning,le2014distributed,mesnil2014ensemble} to split them evenly for training and testing, respectively. 
PV-DBOW represents every movie review using a fixed-length feature vector. It is a binary classification problem. Our implementation benefits from the codes publicly available at \url{https://github.com/mesnilgr/iclr15} and \url{https://github.com/shaform/experiments/tree/master/caffe_sentiment_analysis}.


\subsection{Label Noise}

It is a binary classification problem so that the maximum noise rate that we can generate is $50\%$. 
We test three cases: $r=0.0, 0.2, 0.4$.\\
We choose {{an exponential distribution variant}} as an EDF, i.e., $\lambda=0$. 
Additionally, if $r=0.0$, $\beta>0$, and $\beta<0$ otherwise. 
Specifically:\\
(1) When it is clean, we set $\beta=0.9$ so that larger weights are assigned to harder examples. \\
(2) When noise exists, for $r=0.2\text{~and~} r= 0.4$, we set $\beta=-0.52$ and $\beta=-0.33$, respectively. Therefore, easier examples have larger weights. 
\\
Note that the settings of $\lambda$ and $\beta$ change a lot because we have only two classes here. 

\subsection{Sample Imbalance}
We use all the positive reviews (12500) and randomly sample a small proportion of negative reviews: \\
(1) When P-N Ratio is 10:1, 1250 negative reviews are sampled. \\
(2) When P-N Ratio is 50:1, 250 negative reviews are sampled. 
\\
We set $\lambda=0.3, \beta=6.5$.

\subsection{Net Architecture and Optimisation Solver}

\begin{figure}[!h]
	\centering
	\captionsetup{width=\textwidth}
	\includegraphics[width=0.25\textwidth]{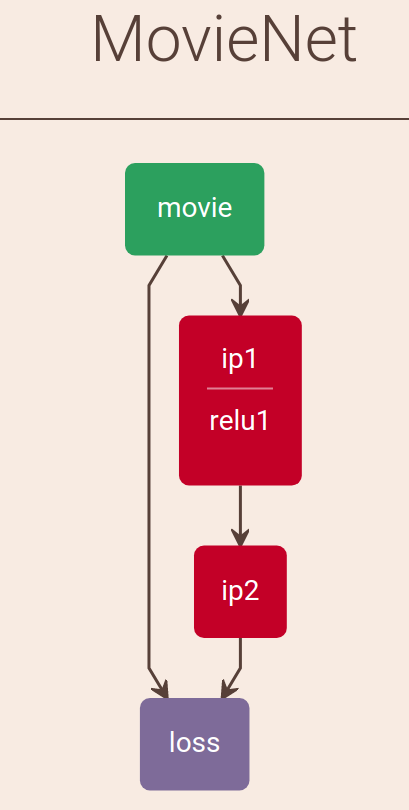}
	\caption{A designed neural network with one 8-neuron hidden layer for sentiment analysis on IMDB. 
		It is quite simple and our objective is not to represent the state-of-the-art.
	}
	\label{fig:IMDB_network}
\end{figure}

The network architecture is shown in Figure~\ref{fig:IMDB_network} and its SGD solver is as follows:

\begin{lstlisting}
# configuration of solver.prototxt 
net: "nn.prototxt"

# test_iter specifies how many forward passes the test should carry out.
# We have test batch size 250 and 100 test iterations,
# covering the full 25,000 testing reviews.
test_iter: 100
# Carry out testing every 500 training iterations.
test_interval: 500

# The base learning rate, momentum and the weight decay of the network.
base_lr: 0.01
momentum: 0.9
weight_decay: 0.002

# The learning rate policy
lr_policy: "inv"
gamma: 0.00001
power: 0.75

# Display every 100 iterations
display: 100
# The maximum number of iterations
max_iter: 10000
snapshot_prefix: "nn"
# solver mode: CPU or GPU
solver_mode: GPU

random_seed: 123
\end{lstlisting}

%

\section{The Effectiveness of Label Correction}
Is it feasible to correct the labels of noisy training data? 
In label correction, we replace the original labels by their corresponding predictions at the end of training. 
The results are shown in Table~\ref{table:CIFAR_noise_three_subsets}.

\begin{table}[!h]
	\caption{
		Is it feasible to correct the labels of noisy training data? 
		Our results demonstrate the effectiveness of label correction using DM. 
		When retraining from scratch on the relabelled training data, we do not adjust the hyper-parameters $\beta$ and $\lambda$. Therefore, the reported results of retraining on relabelled datasets are not the optimal. 
		In label correction, the original labels are replaced by their corresponding predictions.  
	}
	\label{table:CIFAR_noise_three_subsets}
	\centering
	\setlength{\tabcolsep}{4.0pt} 
	\small
	\begin{tabular}{lcccccccccc}
		\toprule
		\multirow{2}{*}{\makecell{Noise\\Rate $r$}}& \multirow{2}{*}{\makecell{Emphasis\\Mode}} & \multirow{2}{*}{Model} & \multicolumn{2}{c}{\makecell{Testing \\Accuracy (\%)}} & \multicolumn{2}{c}{ \makecell{Accuracy  on\\ Training Sets (\%)} }&\multicolumn{2}{c}{ \makecell{Fitting degree \\ of subsets (\%)} } &  & \makecell{Retrain \\after }\\
		\cmidrule{4-5}  \cmidrule{6-7} \cmidrule{8-9}
		& & & Best & Final & Noisy & Intact & Clean & Noisy &  & \makecell{label \\correction}\\
		\hline
		\multirow{3}{*}{20\%}
		&0&CCE & 86.5 & 76.8 & {95.7} & 80.6 & \textbf{99.0} & 85.9 &  & --\\
		&\makecell{${1}/{3} (\lambda=0.5$)}&DM ($\beta=12$) & \textbf{89.4} & \textbf{87.8} & 81.5 & \textbf{95.0} & 98.8 & \textbf{11.7} &  & 89.3 (+1.5)\\
		\midrule \midrule
		\multirow{3}{*}{40\%}
		&0&CCE & 82.8 & 60.9 & {83.0} & 64.4 & \textbf{97.0}& 81.1 &  & -- \\
		&\makecell{ $1/2 (\lambda=1$)}&DM ($\beta=16$) & \textbf{84.7} & \textbf{83.3} & 60.3 & \textbf{88.9} & 94.8& \textbf{7.5} &  & 85.3 (+2)\\
		\bottomrule
	\end{tabular}
\end{table}

\begin{table}[!h]
	\caption{
		Results of CCE, DM on CIFAR-10 under noisy labels. 
		For every model, we show its best test accuracy during training and the final test accuracy when training terminates, 
		which are indicated by `Best' and `Final', respectively. 
		We also present the results on corrupted training sets and original intact one.  The overlap rate between corrupted and intact sets is $(1-r)$. 
		When $\lambda$ is larger, $\beta$ should be larger for adjusting emphasis variance. 
		The intact training set serves as an indicator of meaningful fitting and we observe that its accuracy is always consistent with the final test accuracy.
	}
	\label{table:CIFAR_noise_three}
	\centering
	\setlength{\tabcolsep}{6pt} 
	\fontsize{8.8pt}{9.8pt}\selectfont
	\begin{tabular}{lcccccc}
		\toprule
		\multirow{2}{*}{\makecell{Noise\\Rate $r$}}& \multirow{2}{*}{\makecell{Emphasis\\Mode}} & \multirow{2}{*}{Model} & \multicolumn{2}{c}{\makecell{Testing \\Accuracy (\%)}} & \multicolumn{2}{c}{ \makecell{Accuracy  on\\ Training Sets (\%)} }\\
		\cmidrule{4-5}  \cmidrule{6-7}
		& & & Best & Final & Corrupted & Intact (Meaningful Fitting) \\
		\hline
		\multirow{7}{*}{20\%}
		&0&CCE & 86.5 & 76.8 & \textbf{95.7} & 80.6\\
		\cmidrule{2-7}
		&None & DM ($\beta$=$0$) & 83.5 & 58.1 & 50.6 & 60.2\\
		&\multirow{1}{*}{0 ($\lambda=0$)}&DM ($\beta=2$) & 84.9 & 76.4 & 85.3 & 80.5\\
		&${1}/{3} (\lambda=0.5$)&DM ($\beta=12$) & \textbf{89.4} & \textbf{87.8} & 81.5 & \textbf{95.0}\\
		&${1}/{2} (\lambda=1$) & \makecell{DM ($\beta=16$) } & 87.3 & 86.7 & 78.4 & 93.8\\
		&${2}/{3} (\lambda=2$) &DM ($\beta=24$) & 85.8 & 85.5 & 76.0 & 91.4\\
		\hline \hline
		\multirow{7}{*}{40\%}
		&0&CCE & 82.8 & 60.9 & \textbf{83.0} & 64.4\\
		\cmidrule{2-7}
		&None  & DM ($\beta$=$0$) & 71.8 & 44.9 & 31.3 & 45.8\\
		&\multirow{1}{*}{0 ($\lambda=0$)}&DM ($\beta=1$) & 78.4 & 65.6 & 63.3 & 66.6\\
		&${1}/{3} (\lambda=0.5$) &DM ($\beta=12$) & \textbf{85.1} & {79.9} & 67.7 & {85.7}\\
		&${1}/{2} (\lambda=1$)&DM ($\beta=16$) & 84.7 & \textbf{83.3} & 60.3 & \textbf{88.9}\\
		&${2}/{3} (\lambda=2$)&DM ($\beta=20$) & 52.7 & 52.7 & 35.4 & 53.6\\
		\hline \hline
		\multirow{7}{*}{60\%}
		&0&CCE & 69.5 & 37.2 & \textbf{84.1} & 40.5\\
		\cmidrule{2-7}
		&None & DM ($\beta$=$0$) & 69.9 & 57.9 & 40.1 & 58.6\\
		&\multirow{1}{*}{0 ($\lambda=0$)}&DM ($\beta=0.5$) & 72.3 & 53.9 & 42.1 & 55.1\\
		&${1}/{3} (\lambda=0.5$)&DM ($\beta=12$) & {77.5} & {58.5} & 55.5 & {62.6}\\
		&${1}/{2} (\lambda=1$)&DM ($\beta=12$) & 71.9 & {70.0} & 41.0 & {73.9}\\
		&${2}/{3} (\lambda=2$)&DM ($\beta=12$) & \textbf{80.2} & \textbf{72.5} & 44.9 & \textbf{75.4}\\
		\hline\hline
		\multirow{7}{*}{80\%}
		&0&CCE & 36.1 & 16.1 & \textbf{54.3} & 18.4\\
		\cmidrule{2-7}
		&None & DM ($\beta$=$0$) & 44.4 & 28.2 & 20.6 & 28.8\\
		&\multirow{1}{*}{0 ($\lambda=0$)}&DM ($\beta=0.5$) & 46.2 & 21.3 & 27.8 & 23.1\\
		&${1}/{3} (\lambda=0.5$)&DM ($\beta=8$) & \textbf{51.6} & 22.4 & 46.1 & 24.4\\
		&${1}/{2} (\lambda=1$)&DM ($\beta=8$) & 35.5 & {31.5} & 19.8 & {32.3}\\
		&${2}/{3} (\lambda=2$)&DM ($\beta=12$) & 33.0 & \textbf{32.8} & 14.2 & \textbf{32.6}\\
		\bottomrule
	\end{tabular}
\end{table}

\section{More Detailed Empirical Results}
\label{appendix_sec: ablation_study_onModeAndVariance}

\subsection{Training DNNs under label noise}

\textbf{Practical research question}: \textit{What training examples should be focused on and how much more should they be emphasised when training DNNs under label noise?}

\textbf{Our proposal}: DM incorporates emphasis mode and variance, and serves as explicit regularisation in terms of example weighting.  

\textbf{Important finding}: When label noise rate is higher, we can improve a model's robustness by moving emphasis mode towards relatively less difficult examples.  

\vspace{-0.1cm}
\subsection{Empirical Analysis of DM on CIFAR-10}
\label{sec:empirical_analysis}
\vspace{-0.1cm}

To empirically understand DM well, we explore the behaviours of DM on CIFAR-10 with $r=20\%, 40\%, 60\%, 80\%$, respectively. We use ResNet-56 which has larger capacity than ResNet-20.

\textbf{Design choices.} We mainly analyse the impact of different emphasis modes for different noise rates. We explore five emphasis modes: 
1) None: $\beta=0$. There is no emphasis mode since all examples are treated equally; 
2) 0: $\lambda=0$; 
3) $\frac{1}{3}$: $\lambda=0.5$; 
4) $\frac{1}{2}$: $\lambda=1$; 
5) $\frac{2}{3}$: $\lambda=2$.  
We remark that when $\lambda$ is larger, the emphasis mode is higher, leading to relatively easier training data points are emphasised. 
%
\textit{When emphasis mode changes, emphasis variance changes accordingly}. Therefore, to set a proper spread for each emphasis mode, we try four emphasis variances and choose the best one\footnote{Since there is a large interval between different $\beta$ in our four trials, we deduce that the chosen one is not the optimal.
	The focus of this work is not to optimize the hyper-parameters. 
	Instead, we focus more on the practical research question: \textit{What training examples should be focused on and how much more should they be emphasised when training DNNs under label noise?}
}
to study the impact of emphasis mode.

\begin{figure*}[t!]
	\centering
	\begin{subfigure}[t!]{0.3\textwidth}
		\centering
		\captionsetup{width=1\textwidth}
		\includegraphics[width=\textwidth]{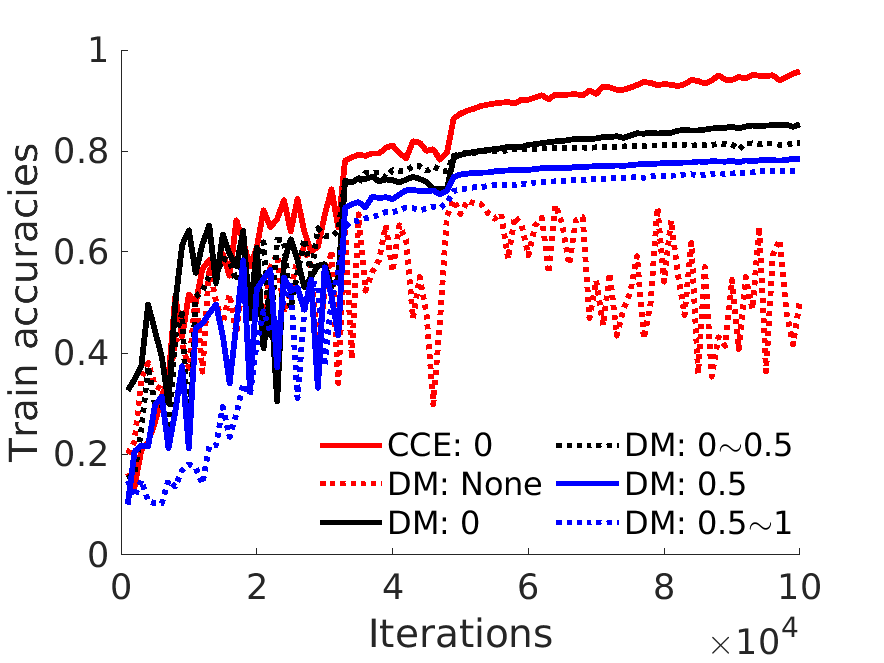}
		\label{fig:N02_Best_trains}
		%
	\end{subfigure}%
	\begin{subfigure}[t!]{0.3\textwidth}
		\centering
		\captionsetup{width=1\textwidth}
		\includegraphics[width=\textwidth]{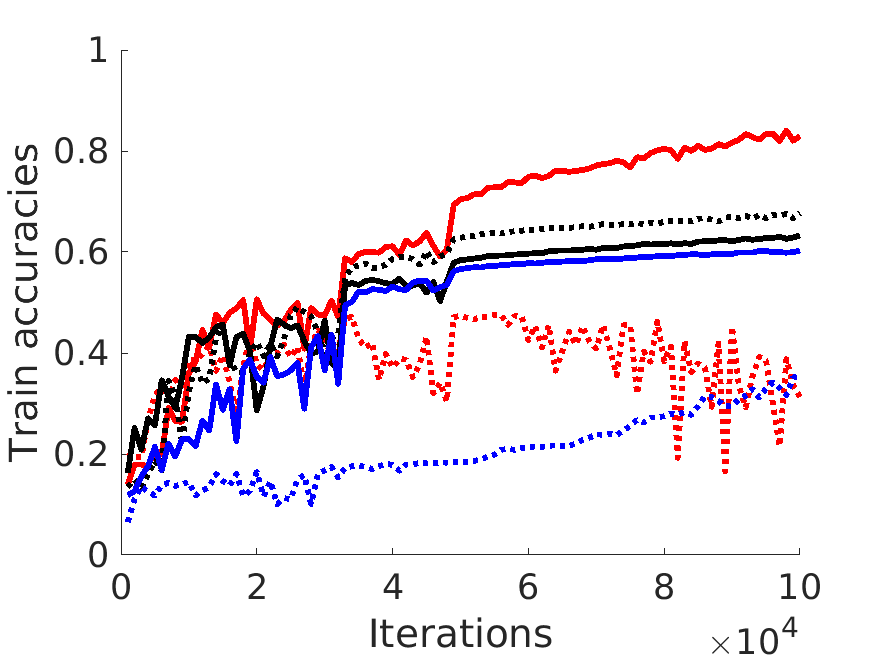}
		\label{fig:N04_Best_trains}
		%
	\end{subfigure}
	\begin{subfigure}[t!]{0.3\textwidth}
		\centering
		\captionsetup{width=1\textwidth}
		\includegraphics[width=\textwidth]{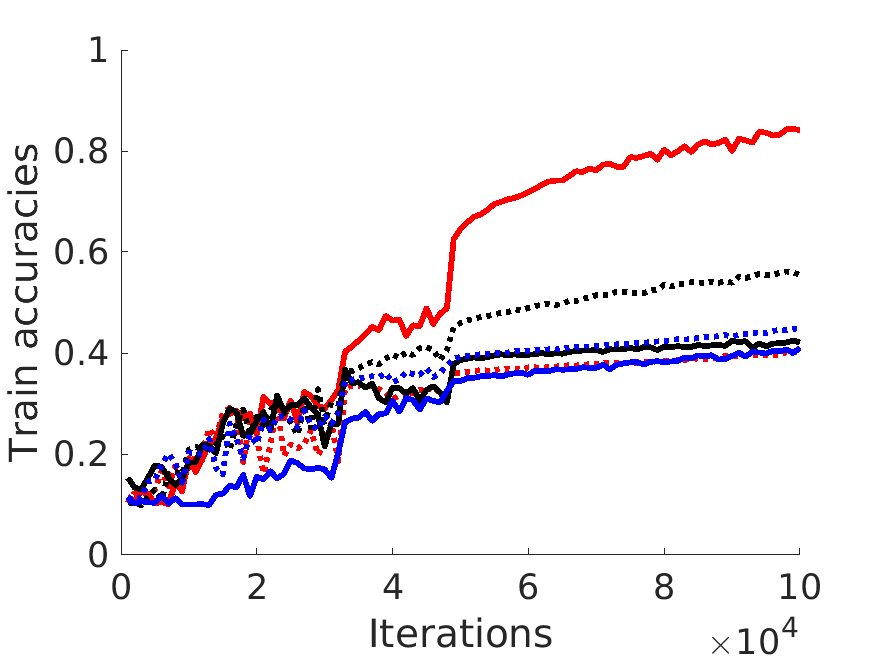}
		\label{fig:N06_Best_trains}
		%
	\end{subfigure}%
	\vspace{-0.45cm}
	
	\begin{subfigure}[t!]{0.3\textwidth}
		\centering
		\captionsetup{width=1\textwidth}
		\includegraphics[width=\textwidth]{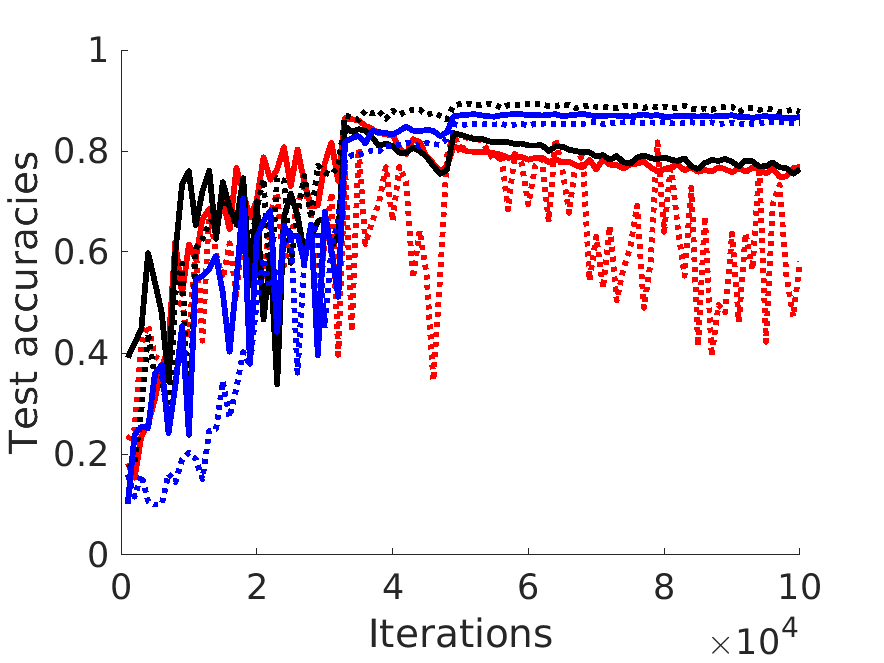}
		\caption{$r=20\%.$}
		\label{fig:N02_Best_tests}
	\end{subfigure}
	\begin{subfigure}[t!]{0.3\textwidth}
		\centering
		\captionsetup{width=1\textwidth}
		\includegraphics[width=\textwidth]{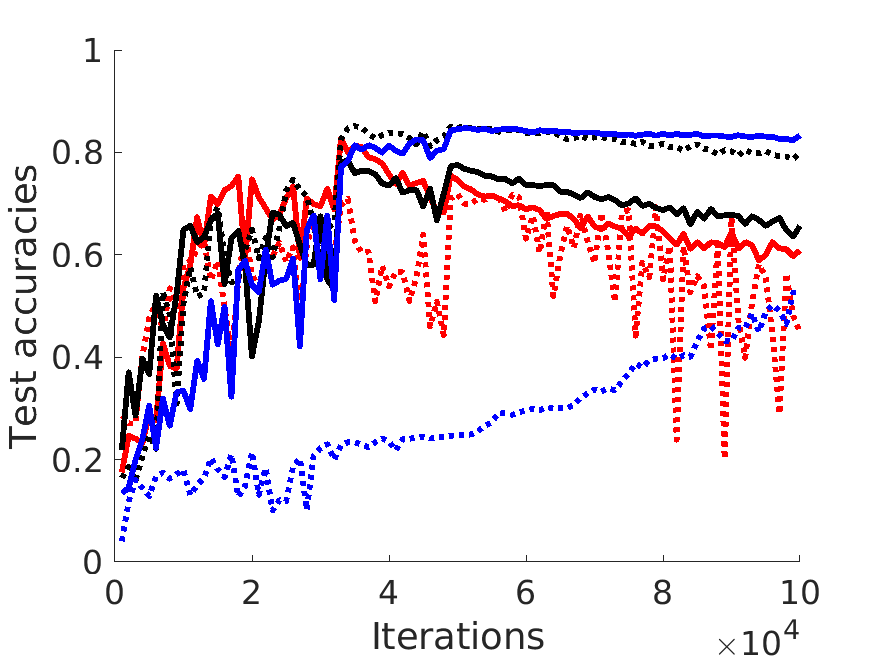}
		\caption{$r=40\%.$}
		\label{fig:N04_Best_tests}
	\end{subfigure}%
	\begin{subfigure}[t!]{0.3\textwidth}
		\centering
		\captionsetup{width=1\textwidth}
		\includegraphics[width=\textwidth]{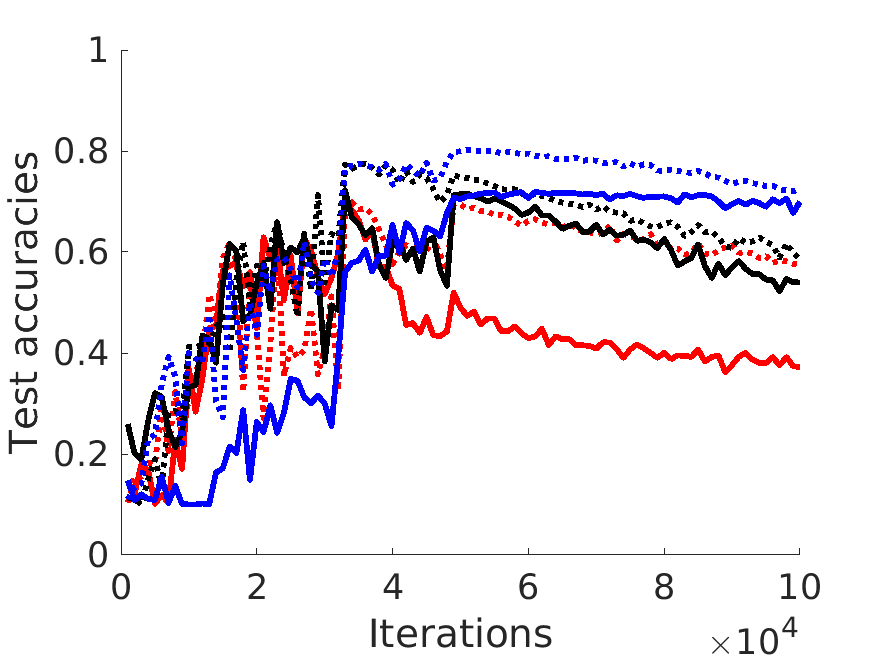}
		\caption{$r=60\%.$}
		\label{fig:N06_Best_tests}
	\end{subfigure}
	%
	%
	\caption{The learning dynamics of ResNet-56 on CIFAR-10, i.e., training and testing accuracies along with training iterations.
		The legend in the top left is shared by all subfigures.  `xxx: yyy' means `method: emphasis mode'. 
		We have two key observations: 1) When noise rate increases, better generalisation is obtained with higher emphasis mode, i.e., focusing on relatively easier examples;
		2) Both overfitting and underfitting lead to bad generalisation. For example, `CCE: 0' fits training data much better than the others while `DM: None' generally fits it unstably or a lot worse. {Better viewed in colour.}
	}
	\label{fig:N02_N04_N06_Bests}
\end{figure*}

\textbf{Results analysis.} We show the results in Table~\ref{table:CIFAR_noise_three}.
The intact training set serves as an indicator of meaningful fitting and we observe that its accuracy is always consistent with the final test accuracy.
We display the training dynamics in Figure~\ref{fig:N02_N04_N06_Bests}. 
We summarise our observations as follows:

\vspace{-0.1cm}
\textit{Fitting and generalisation.}
We observe that CCE always achieves the best accuracy on corrupted training sets, which indicates that CCE has a strong data fitting ability 
even if there is severe noise \citep{zhang2017understanding}. 
As a result, CCE has much worse final test accuracy than most models.   

\vspace{-0.1cm}
\textit{Emphasising on harder examples.} When there exists label noise, we obtain the worst 
final test accuracy
if emphasis mode is 0, i.e., CCE and DM with $\lambda=0$. 
This unveils that in applications where we have  to learn from noisy training data, it will hurt the model's generalisation dramatically if we use CCE or simply focus on harder training data points. 

\vspace{-0.05cm}
\textit{Emphasis mode.} When noise rate is 0, 20\%,  40\%, 60\%, and 80\%, we obtain the best final test accuracy when $\lambda=0$, $\lambda=0.5$, $\lambda=1$, $\lambda=2$, and $\lambda=2$, respectively. 
This demonstrates that
{when noise rate is higher, we can improve a model's robustness by moving emphasis mode towards relatively less difficult examples with a larger $\lambda$, }which is informative in practice.   

\vspace{-0.1cm}
\textit{Emphasis spread.}
As displayed in Table~\ref{table:CIFAR_noise_three} and  Figures~\ref{fig:ResNet56_N02}-\ref{fig:ResNet56_N08} in the supplementary material, emphasis variance also matters a lot when fixing emphasis mode, i.e., fixing $\lambda$. 
For example in Table~\ref{table:CIFAR_noise_three} , when $\lambda=0$, although focusing on harder examples similarly with CCE, DM can outperform CCE by modifying the emphasis variance. 
As shown in Figures~\ref{fig:ResNet56_N02}-\ref{fig:ResNet56_N08}, some models even collapse and cannot converge if the emphasis variance is not rational.

\subsection{Learning Dynamics on Clean Training Data}

\begin{figure}[h!]
	\centering
	\begin{subfigure}[t!]{0.4\textwidth}
		\centering
		\captionsetup{width=1\textwidth}
		\includegraphics[width=\textwidth]{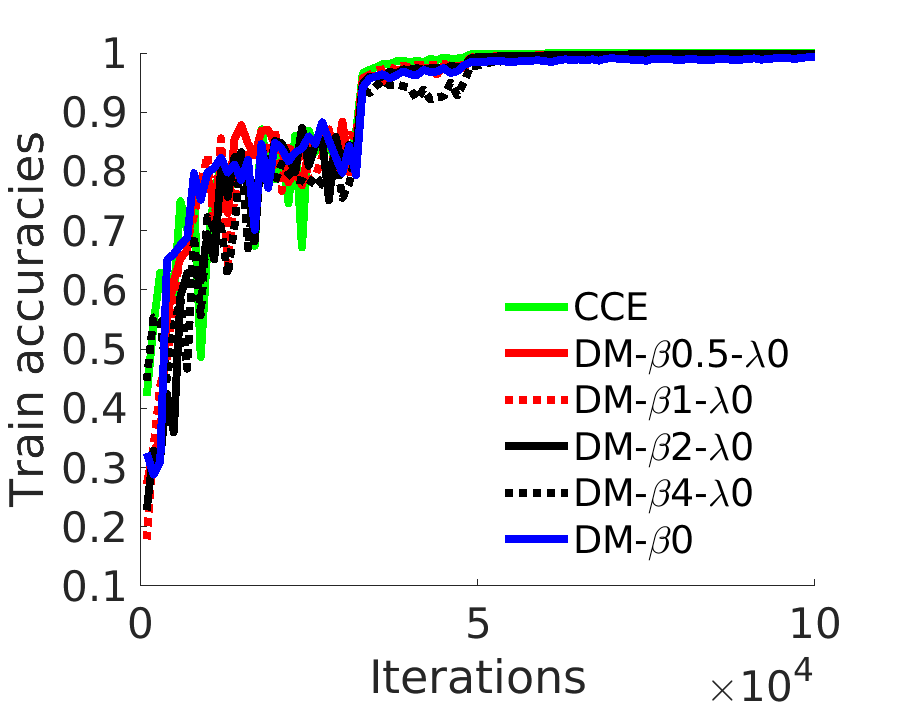}
		\label{fig:training_accuracies}
	\end{subfigure}%
	\begin{subfigure}[t!]{0.4\textwidth}
		\centering
		\captionsetup{width=1\textwidth}
		\includegraphics[width=\textwidth]{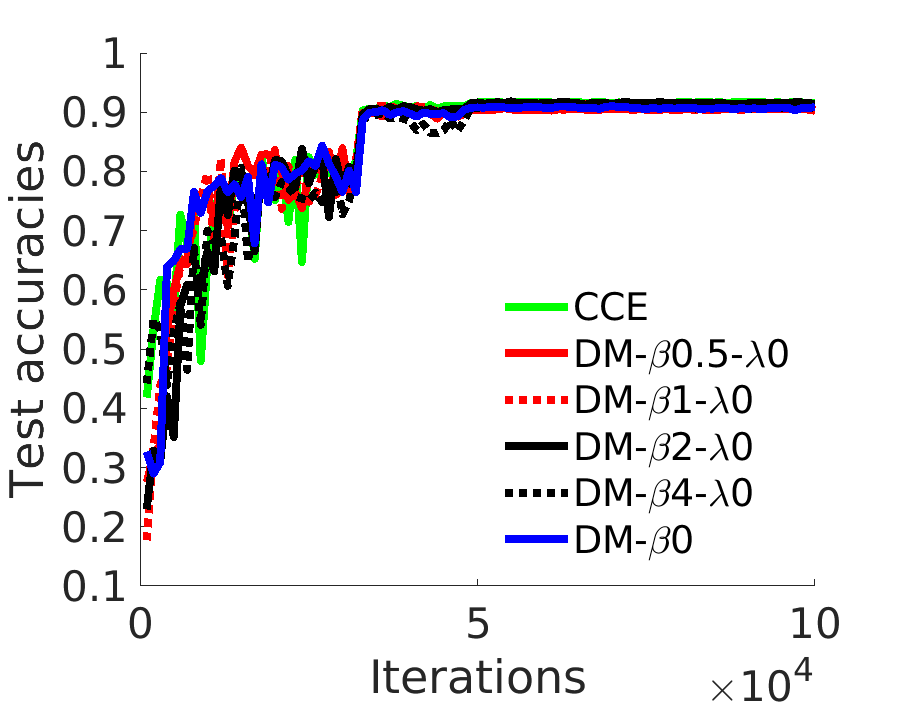}
		\label{fig:training_accuracies}
	\end{subfigure}
	\caption{The training and test accuracies on clean CIFAR-10 along with training iterations.
		The training labels are clean. 
		We fix $\lambda=0$ to focus on harder examples while changing emphasis variance controller $\beta$. 
		\textit{The backbone is ResNet-20.  }
		The results of ResNet-56 are shown in Figure~\ref{fig:clean_cifars_resnet56}.
	}
	\label{fig:clean_cifars_resnet20}
\end{figure}

\begin{figure}[h!]
	\centering
	\begin{subfigure}[t!]{0.4\textwidth}
		\centering
		\captionsetup{width=1\textwidth}
		\includegraphics[width=\textwidth]{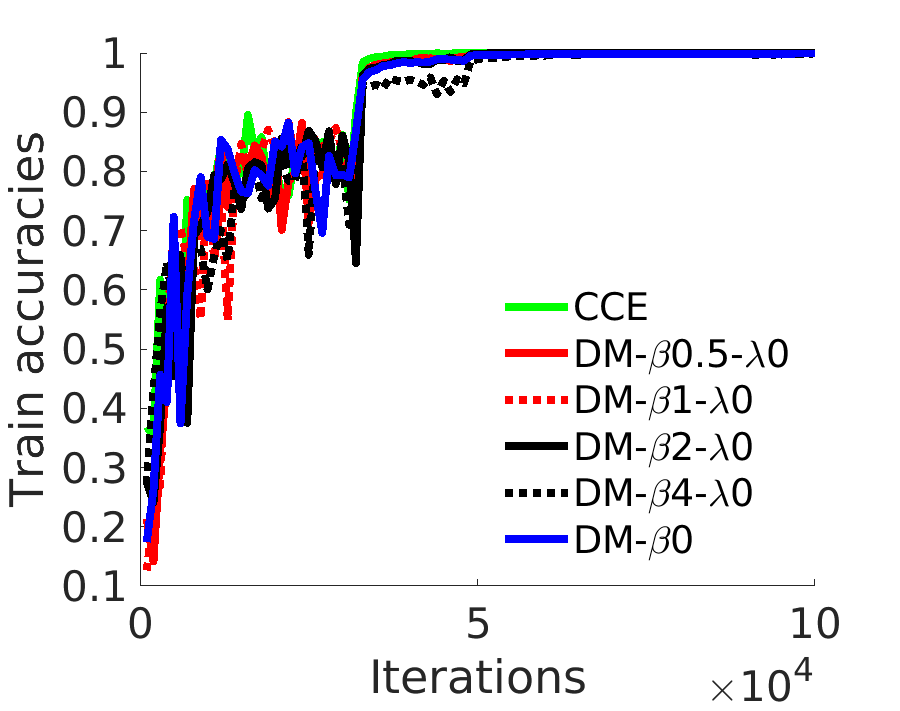}
		\label{fig:Train_ResNet56_N00_CCE_USW_lambda0_and1_different_T_V2}
	\end{subfigure}%
	\begin{subfigure}[t!]{0.4\textwidth}
		\centering
		\captionsetup{width=1\textwidth}
		\includegraphics[width=\textwidth]{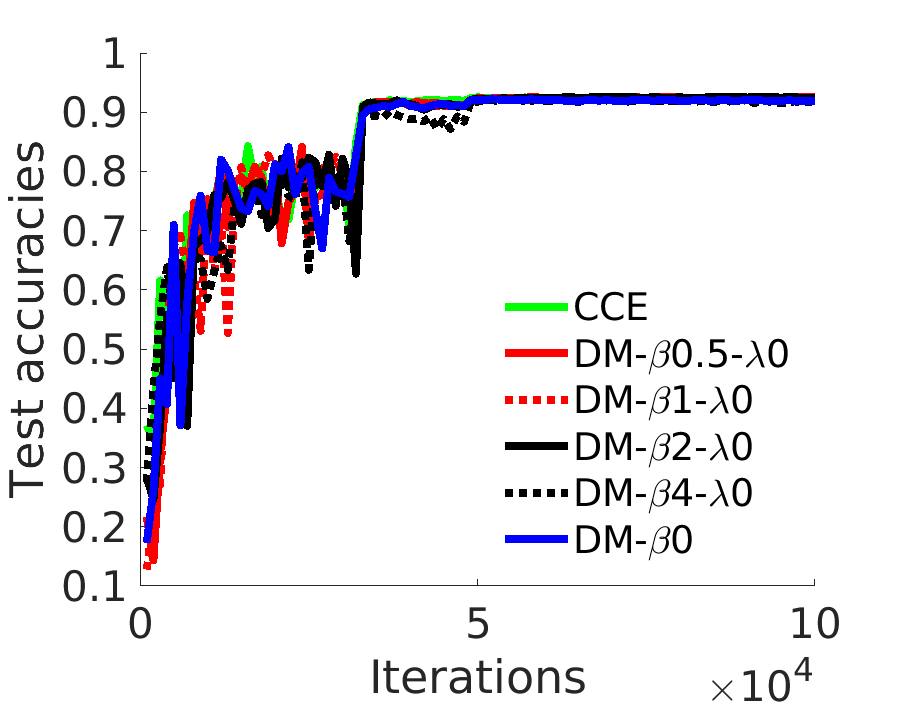}
		\label{fig:Test_ResNet56_N00_CCE_USW_lambda0_and1_different_T_V2}
	\end{subfigure}
	%
	\caption{The training and test accuracies on clean CIFAR-10 along with training iterations.
		The training labels are clean. 
		We fix $\lambda=0$ to focus on more difficult examples while changing emphasis variance controller
		$\beta$. 
		\textit{The backbone is ResNet-56.  }
		The results of ResNet-20 are shown in Figure~\ref{fig:clean_cifars_resnet20}.
	}
	\label{fig:clean_cifars_resnet56}
\end{figure}

The learning dynamics on clean training data are displayed in the Figures~\ref{fig:clean_cifars_resnet20}-\ref{fig:clean_cifars_resnet56}.

%
%

\begin{figure*}[h!]
	\centering
	\begin{subfigure}[t!]{0.245\textwidth}
		\centering
		\captionsetup{width=1\textwidth}
		\includegraphics[width=\textwidth]{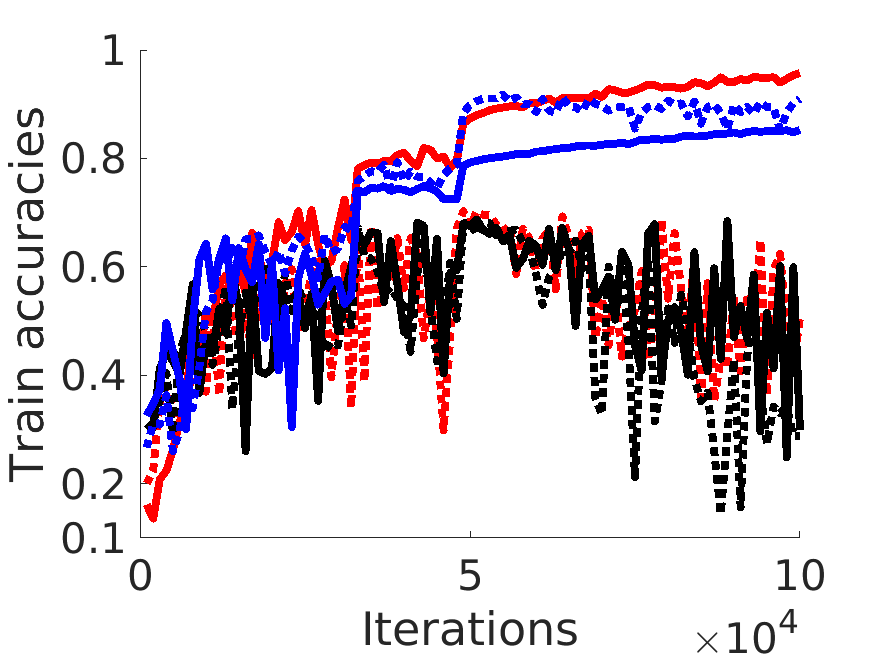}
		\label{fig:training_accuracies}
	\end{subfigure}%
	\begin{subfigure}[t!]{0.245\textwidth}
		\centering
		\captionsetup{width=1\textwidth}
		\includegraphics[width=\textwidth]{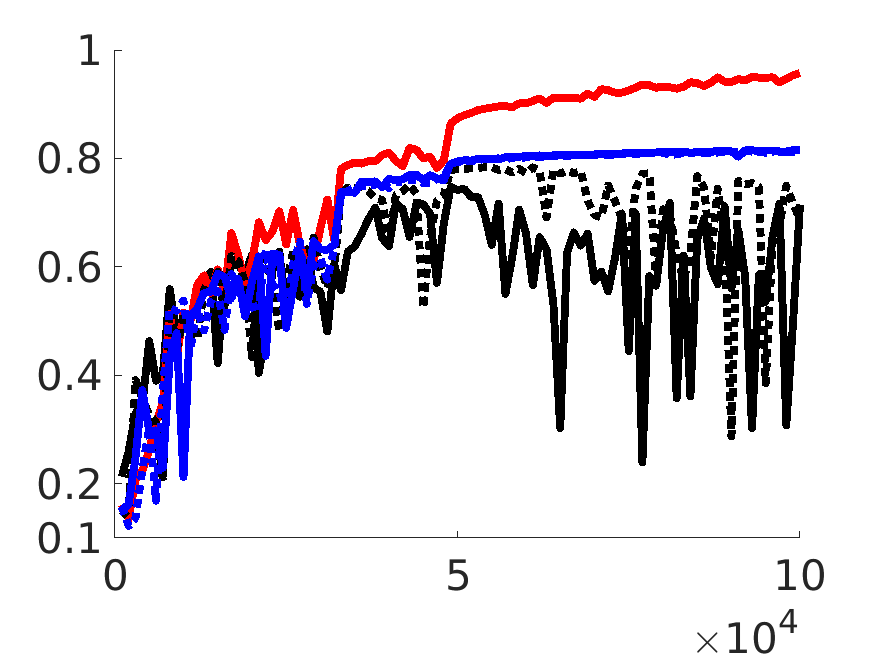}
		\label{fig:training_accuracies}
	\end{subfigure}
	\begin{subfigure}[t!]{0.245\textwidth}
		\centering
		\captionsetup{width=1\textwidth}
		\includegraphics[width=\textwidth]{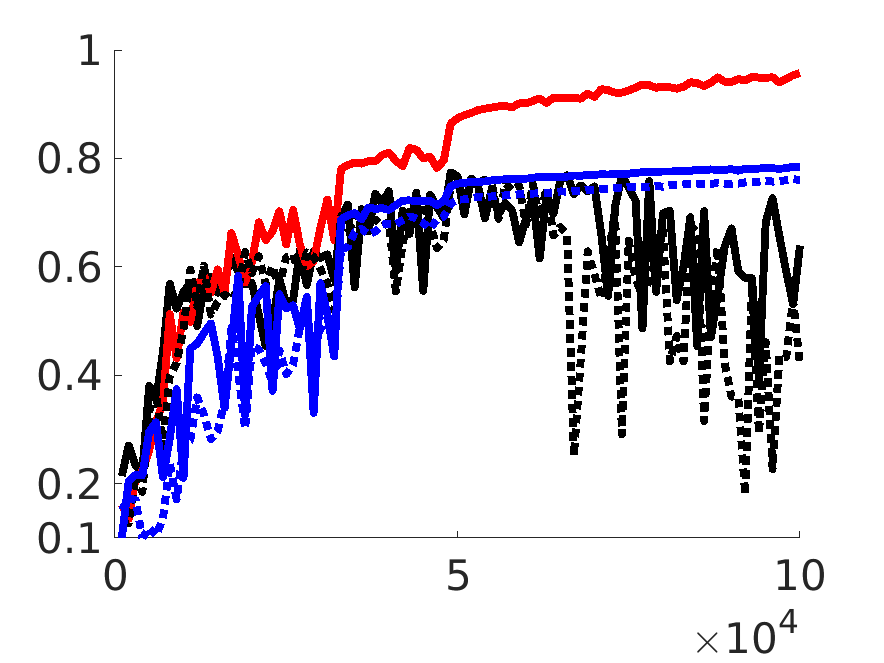}
		\label{fig:training_accuracies}
	\end{subfigure}%
	\begin{subfigure}[t!]{0.245\textwidth}
		\centering
		\captionsetup{width=1\textwidth}
		\includegraphics[width=\textwidth]{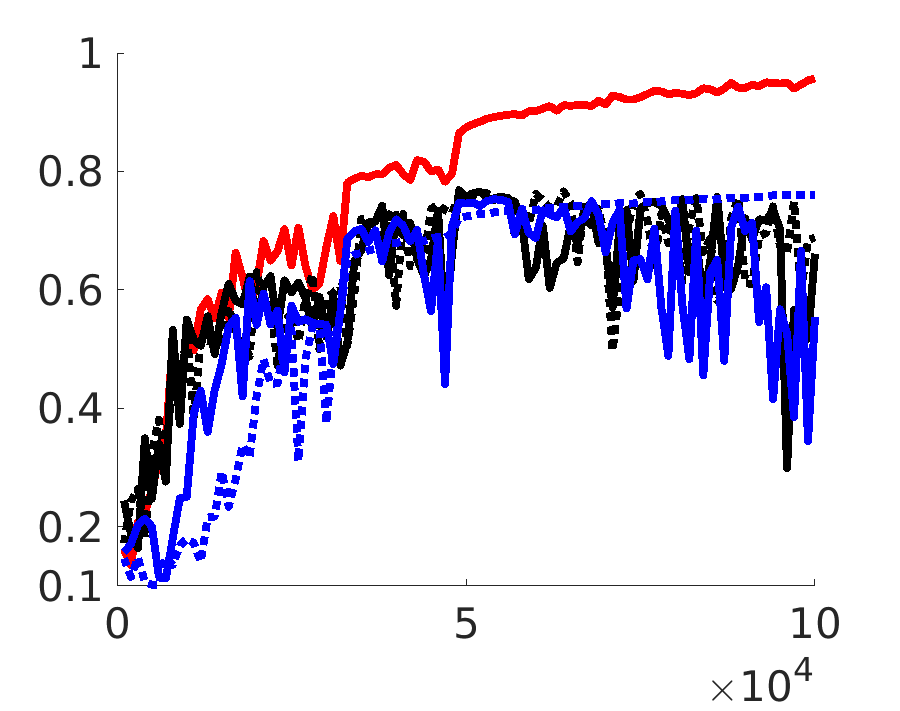}
		\label{fig:training_accuracies}
	\end{subfigure}
	
	\begin{subfigure}[t!]{0.245\textwidth}
		\centering
		\captionsetup{width=1\textwidth}
		\includegraphics[width=\textwidth]{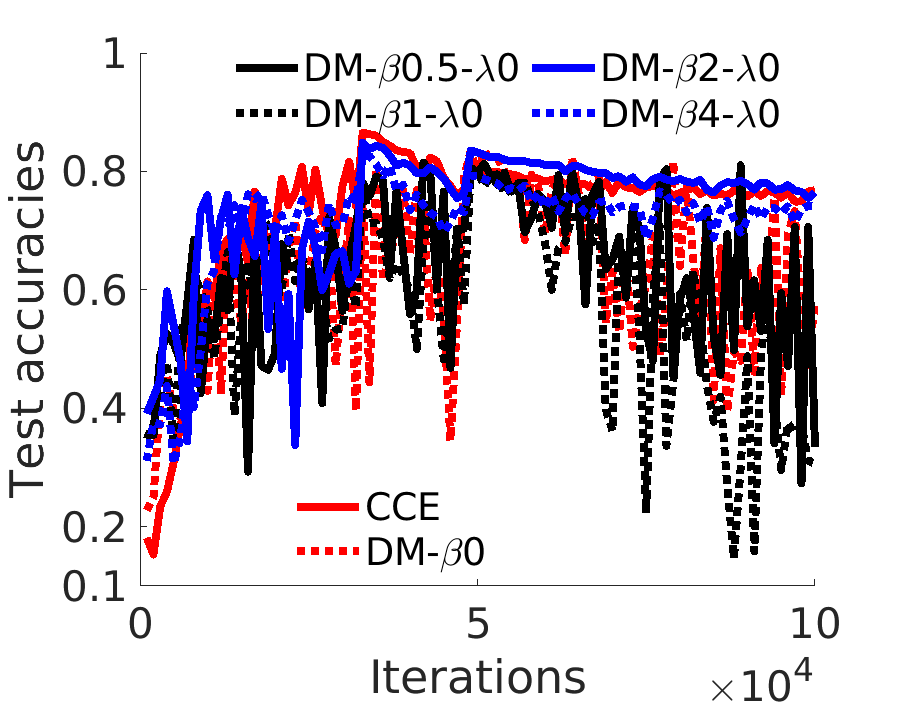}
		\label{fig:training_accuracies}
		%
	\end{subfigure}%
	\begin{subfigure}[t!]{0.245\textwidth}
		\centering
		\captionsetup{width=1\textwidth}
		\includegraphics[width=\textwidth]{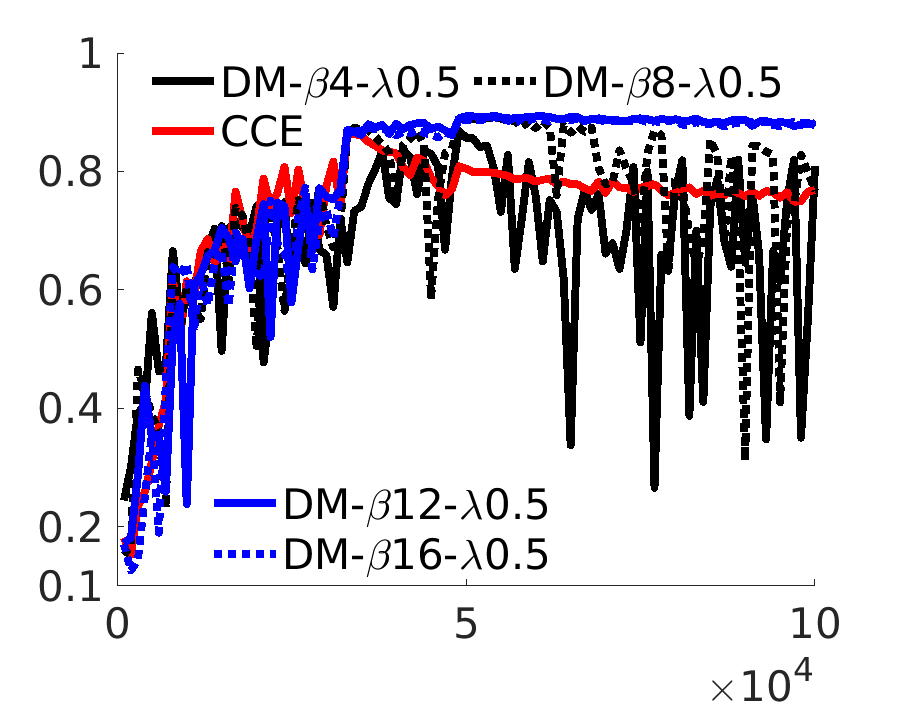}
		\label{fig:training_accuracies}
		%
	\end{subfigure}
	\begin{subfigure}[t!]{0.245\textwidth}
		\centering
		\captionsetup{width=1\textwidth}
		\includegraphics[width=\textwidth]{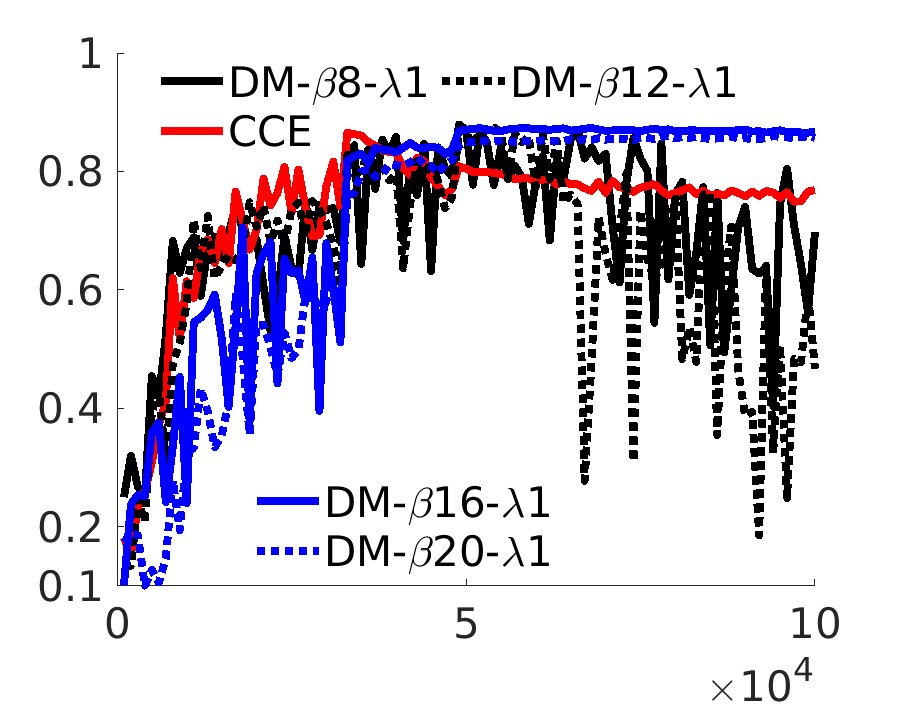}
		\label{fig:training_accuracies}
		%
	\end{subfigure}%
	\begin{subfigure}[t!]{0.245\textwidth}
		\centering
		\captionsetup{width=1\textwidth}
		\includegraphics[width=\textwidth]{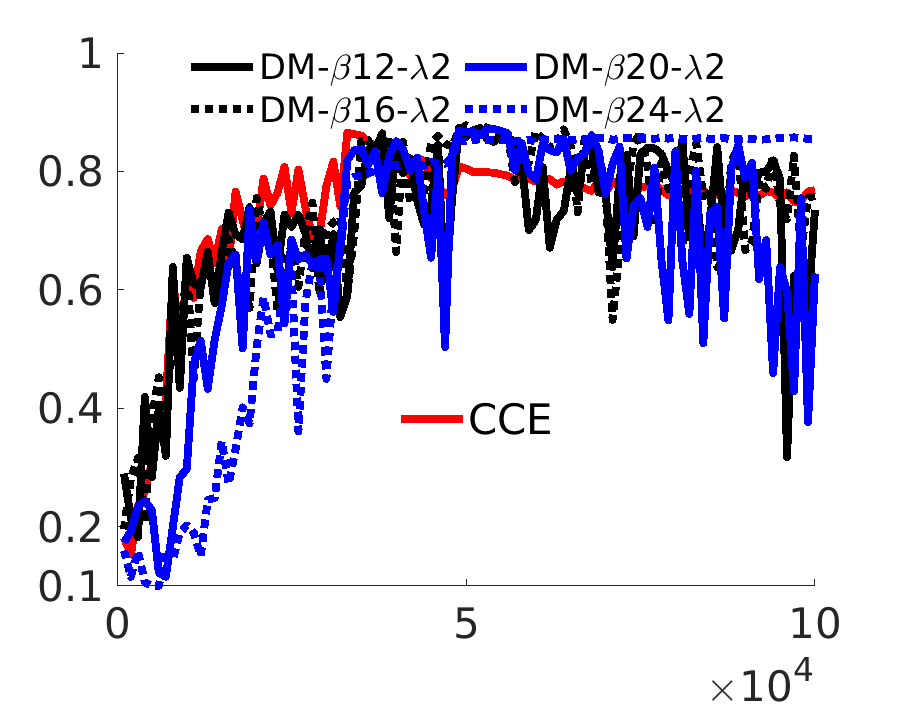}
		\label{fig:training_accuracies}
		%
	\end{subfigure}
	\caption{ResNet-56 on CIFAR-10 ($r=20\%$).
		From left to right, the results of four emphasis modes 0, $\frac{1}{3}$, 0.5, $\frac{2}{3}$ with different emphasis variances are displayed in each column respectively. 
		When $\lambda$ is larger, $\beta$ should be larger.
		Specifically :\\
		1) when $\lambda=0$: we tried $\beta=0.5, 1, 2, 4$;\\ 
		2) when $\lambda=0.5$: we tried $\beta=4, 8, 12, 16$;\\ 
		3) when $\lambda=1$: we tried $\beta=8, 12, 16, 20$; \\
		4) when $\lambda=2$: we tried $\beta=12, 16, 20, 24$. 
	}
	\label{fig:ResNet56_N02}
\end{figure*}

\begin{figure*}[h!]
	\centering
	\begin{subfigure}[t!]{0.245\textwidth}
		\centering
		\captionsetup{width=1\textwidth}
		\includegraphics[width=\textwidth]{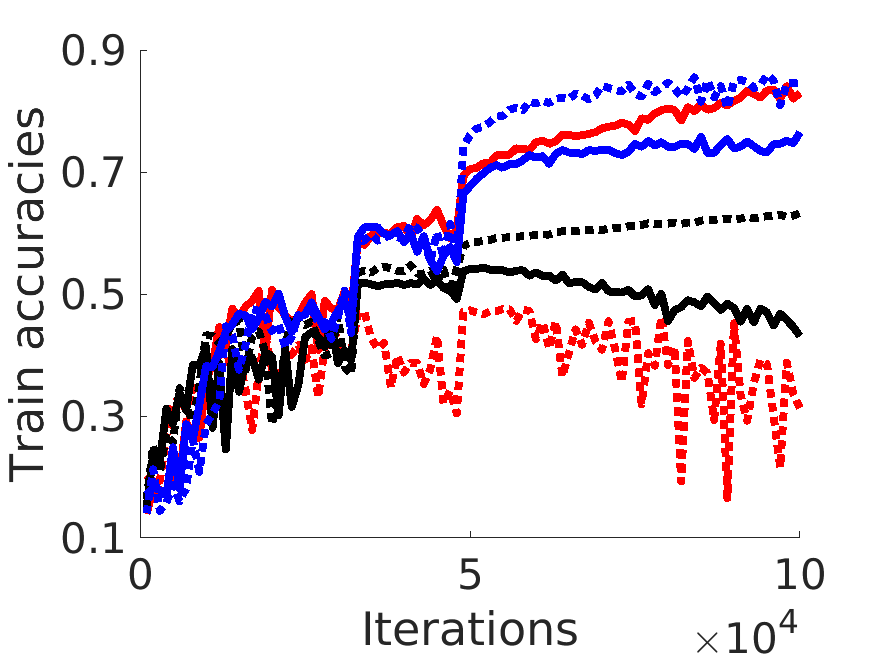}
		\label{fig:training_accuracies}
	\end{subfigure}%
	\begin{subfigure}[t!]{0.245\textwidth}
		\centering
		\captionsetup{width=1\textwidth}
		\includegraphics[width=\textwidth]{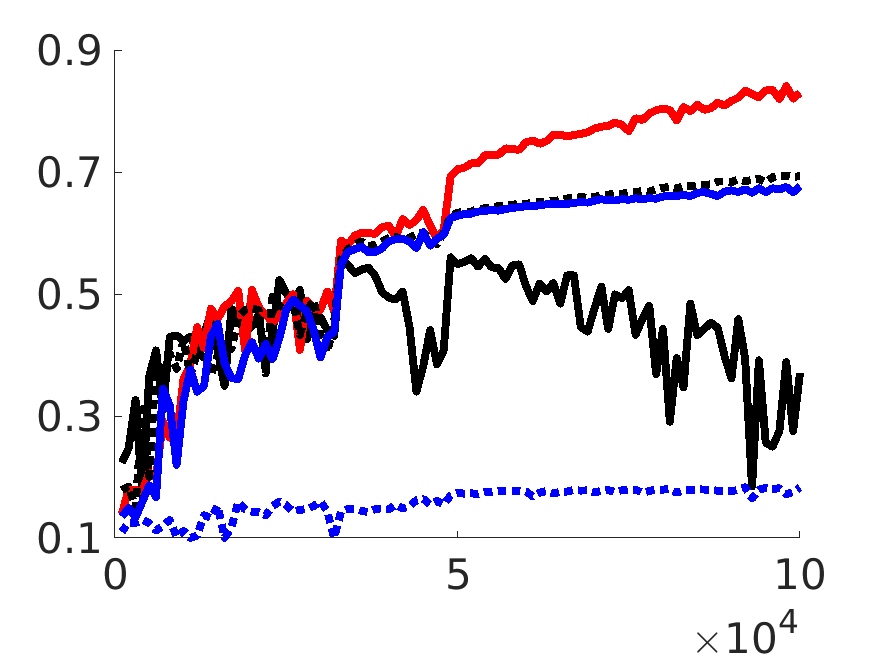}
		\label{fig:training_accuracies}
	\end{subfigure}
	\begin{subfigure}[t!]{0.245\textwidth}
		\centering
		\captionsetup{width=1\textwidth}
		\includegraphics[width=\textwidth]{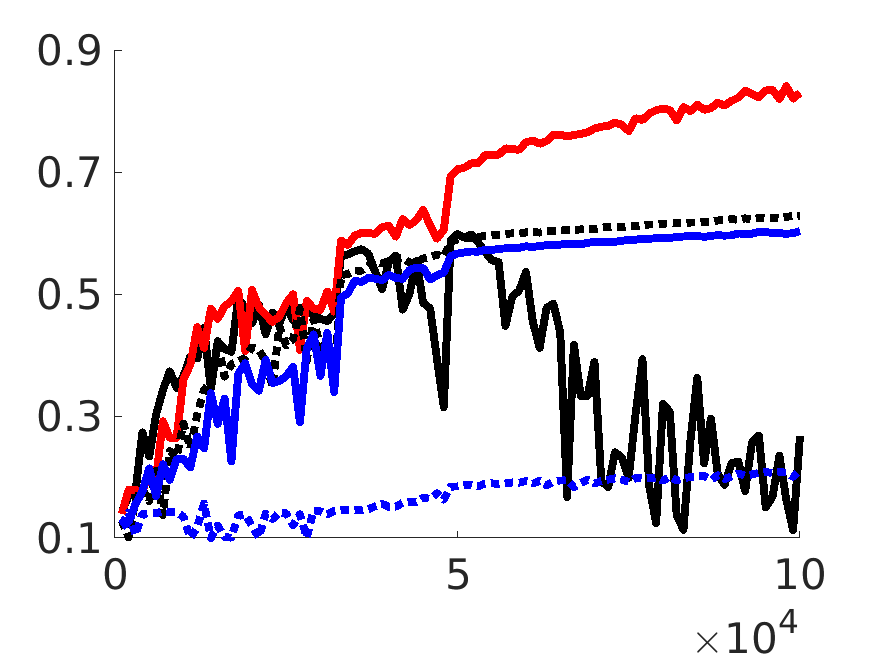}
		\label{fig:training_accuracies}
	\end{subfigure}%
	\begin{subfigure}[t!]{0.245\textwidth}
		\centering
		\captionsetup{width=1\textwidth}
		\includegraphics[width=\textwidth]{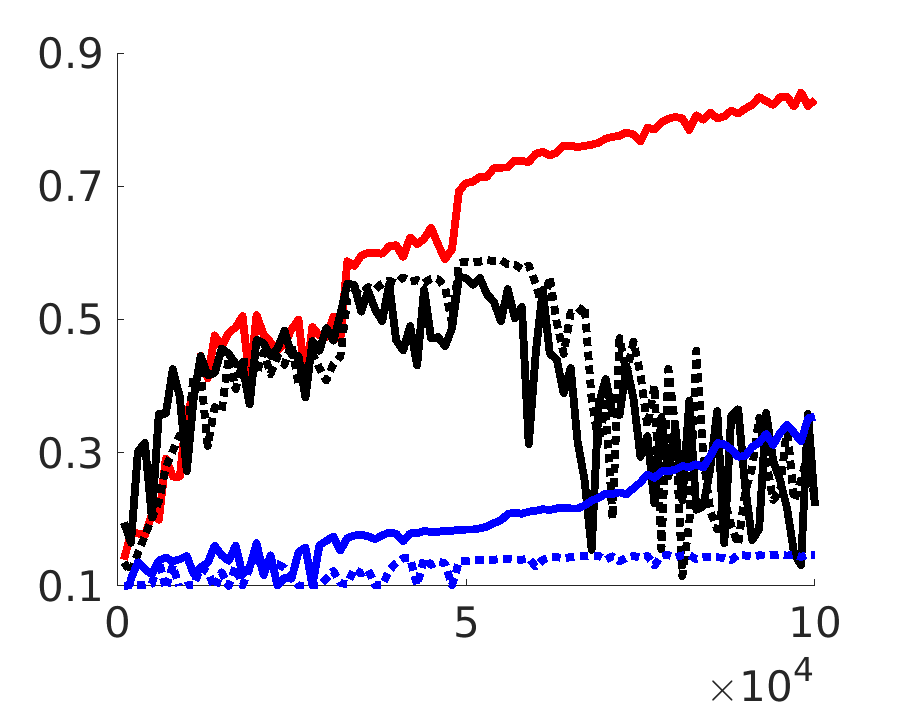}
		\label{fig:training_accuracies}
	\end{subfigure}
	
	\begin{subfigure}[t!]{0.245\textwidth}
		\centering
		\captionsetup{width=1\textwidth}
		\includegraphics[width=\textwidth]{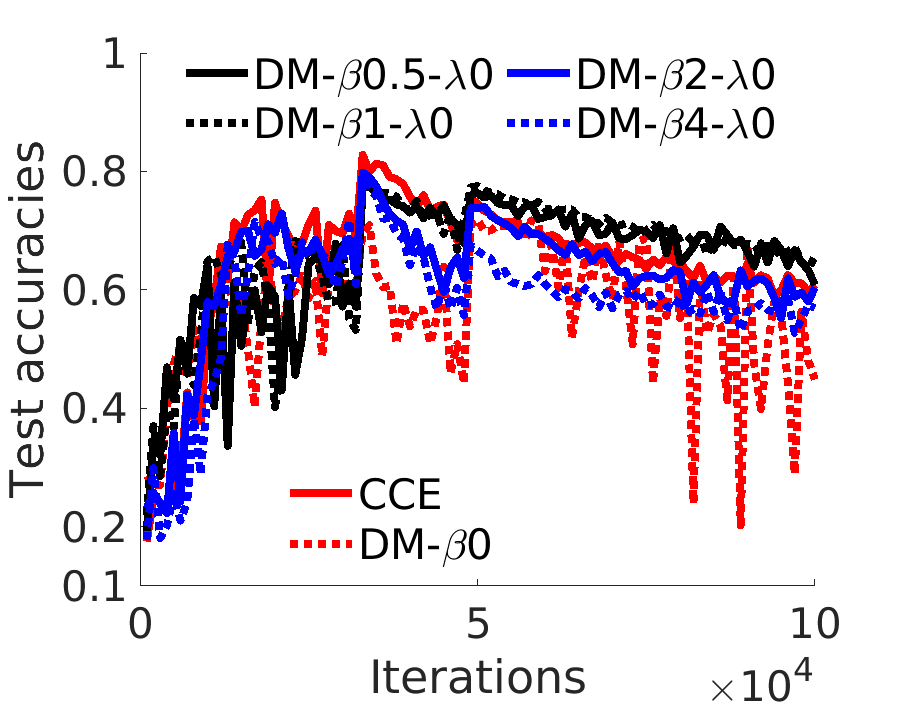}
		\label{fig:training_accuracies}
		%
	\end{subfigure}%
	\begin{subfigure}[t!]{0.245\textwidth}
		\centering
		\captionsetup{width=1\textwidth}
		\includegraphics[width=\textwidth]{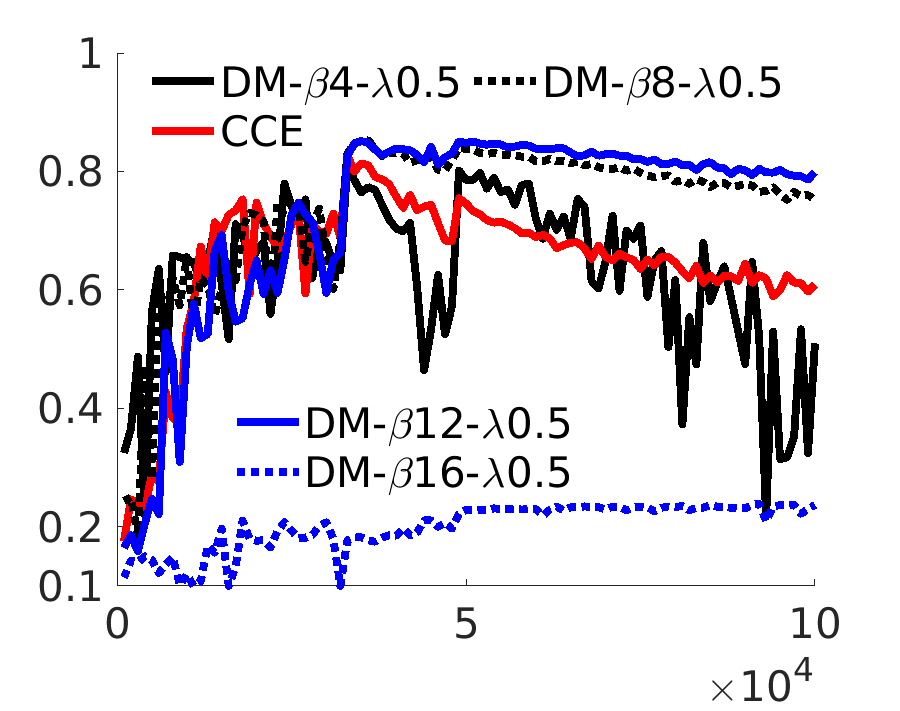}
		\label{fig:training_accuracies}
		%
	\end{subfigure}
	\begin{subfigure}[t!]{0.245\textwidth}
		\centering
		\captionsetup{width=1\textwidth}
		\includegraphics[width=\textwidth]{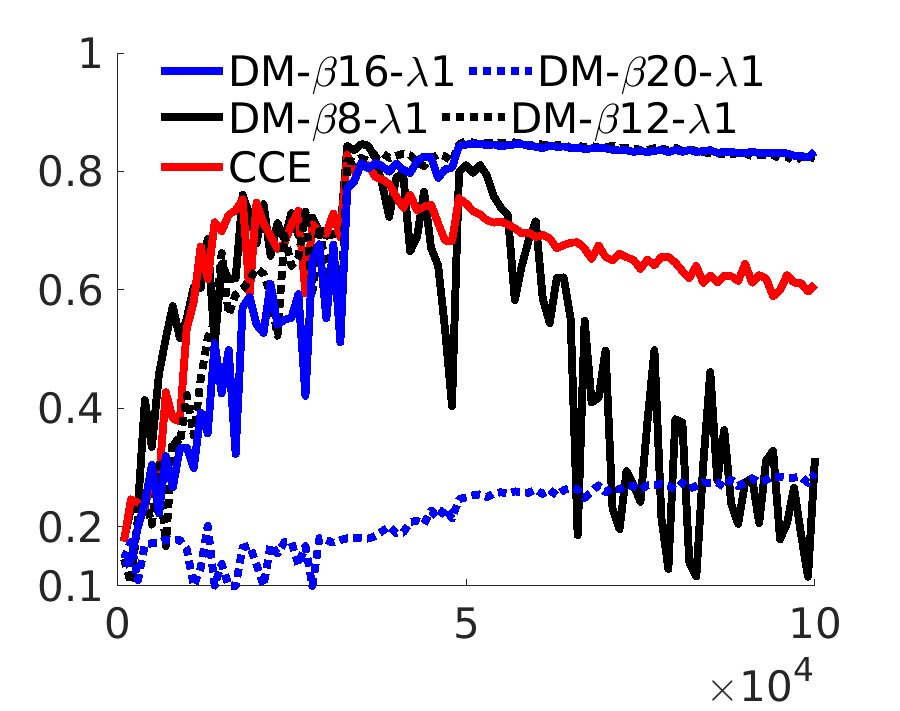}
		\label{fig:training_accuracies}
		%
	\end{subfigure}%
	\begin{subfigure}[t!]{0.245\textwidth}
		\centering
		\captionsetup{width=1\textwidth}
		\includegraphics[width=\textwidth]{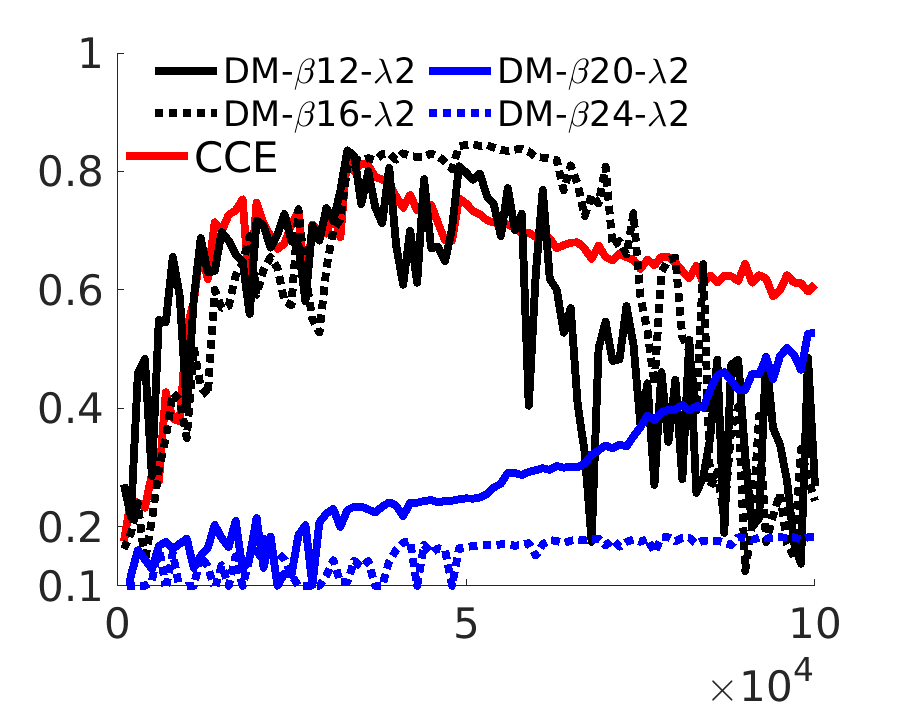}
		\label{fig:training_accuracies}
		%
	\end{subfigure}
	\caption{ResNet-56 on CIFAR-10 ($r=40\%$).
		From left to right, the results of four emphasis modes 0, $\frac{1}{3}$, 0.5, $\frac{2}{3}$ with different emphasis variances are displayed in each column respectively. 
		When $\lambda$ is larger, $\beta$ should be larger.
		Specifically :\\
		1) when $\lambda=0$: we tried $\beta=0.5, 1, 2, 4$;\\ 
		2) when $\lambda=0.5$: we tried $\beta=4, 8, 12, 16$;\\ 
		3) when $\lambda=1$: we tried $\beta=8, 12, 16, 20$; \\
		4) when $\lambda=2$: we tried $\beta=12, 16, 20, 24$.  
	}
	\label{fig:ResNet56_N04}
\end{figure*}

\begin{figure*}[h!]
	\centering
	\begin{subfigure}[t!]{0.245\textwidth}
		\centering
		\captionsetup{width=1\textwidth}
		\includegraphics[width=\textwidth]{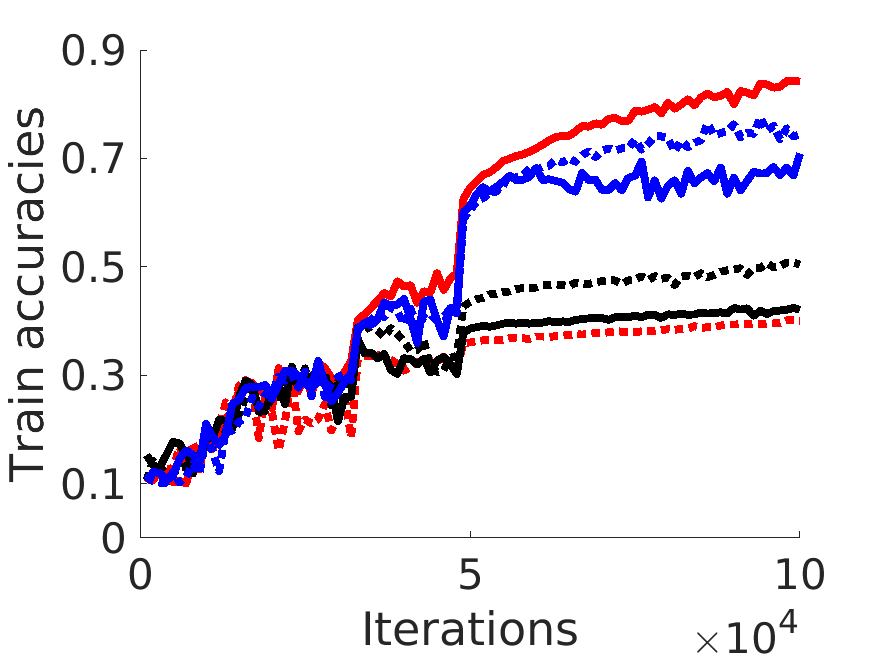}
		\label{fig:training_accuracies}
	\end{subfigure}%
	\begin{subfigure}[t!]{0.245\textwidth}
		\centering
		\captionsetup{width=1\textwidth}
		\includegraphics[width=\textwidth]{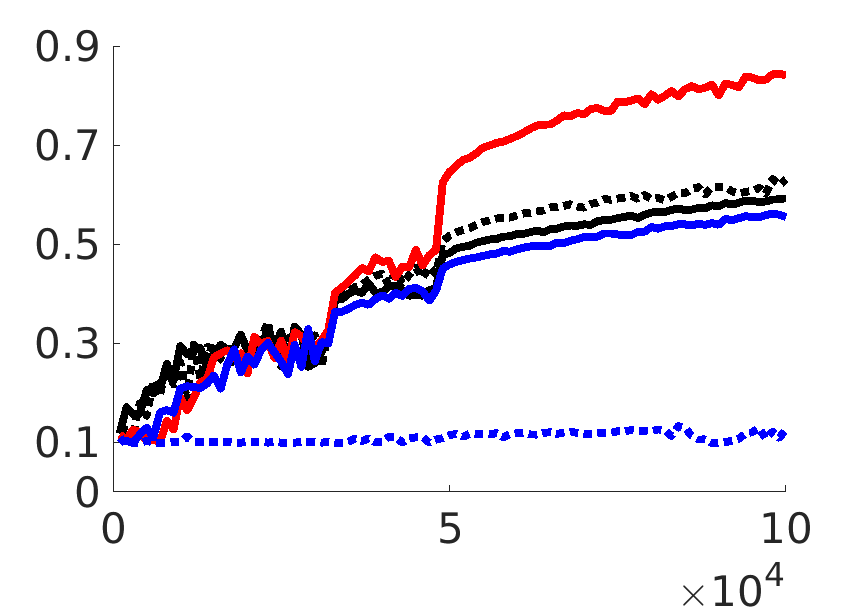}
		\label{fig:training_accuracies}
	\end{subfigure}
	\begin{subfigure}[t!]{0.245\textwidth}
		\centering
		\captionsetup{width=1\textwidth}
		\includegraphics[width=\textwidth]{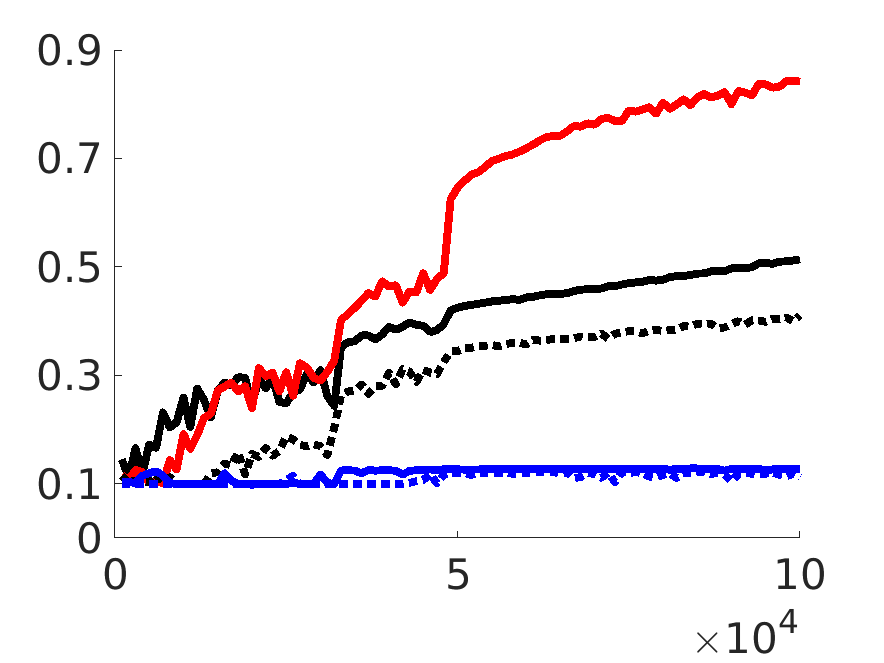}
		\label{fig:training_accuracies}
	\end{subfigure}%
	\begin{subfigure}[t!]{0.245\textwidth}
		\centering
		\captionsetup{width=1\textwidth}
		\includegraphics[width=\textwidth]{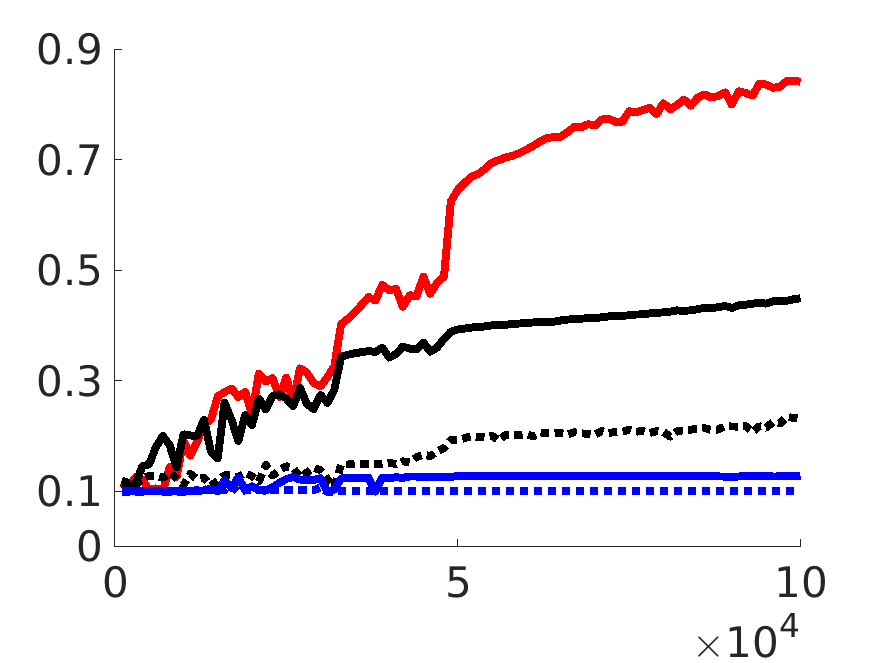}
		\label{fig:training_accuracies}
	\end{subfigure}

	\begin{subfigure}[t!]{0.245\textwidth}
		\centering
		\captionsetup{width=1\textwidth}
		\includegraphics[width=\textwidth]{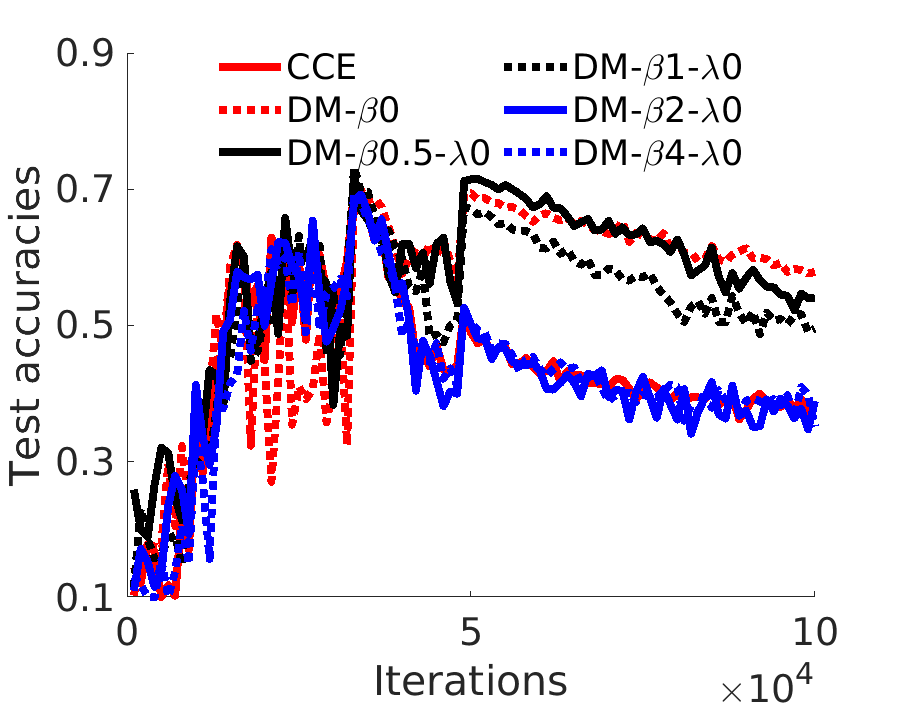}
		\label{fig:training_accuracies}
		%
	\end{subfigure}%
	\begin{subfigure}[t!]{0.245\textwidth}
		\centering
		\captionsetup{width=1\textwidth}
		\includegraphics[width=\textwidth]{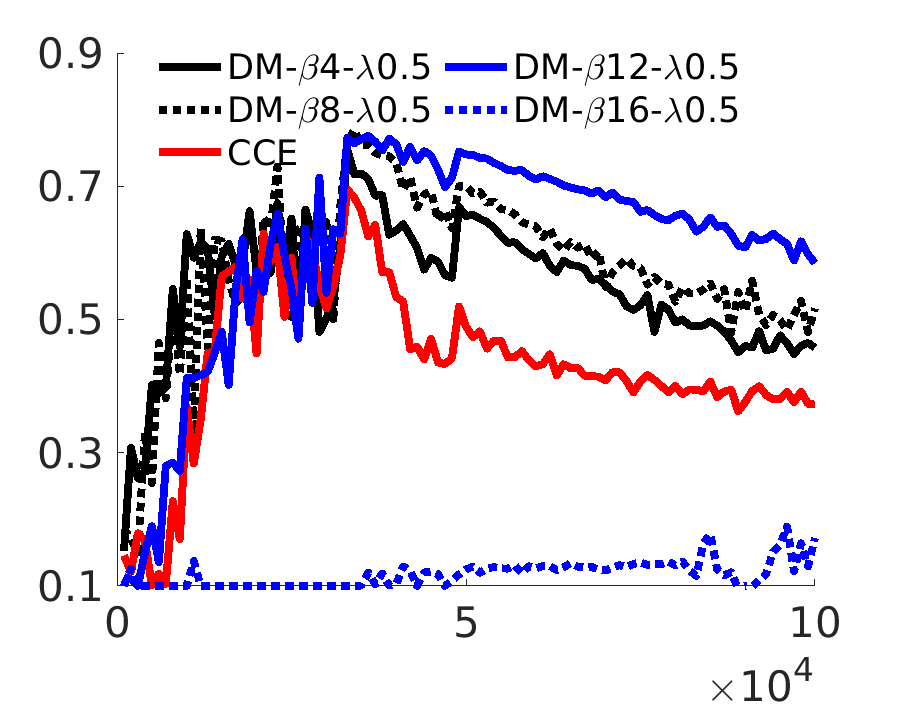}
		\label{fig:training_accuracies}
		%
	\end{subfigure}
	\begin{subfigure}[t!]{0.245\textwidth}
		\centering
		\captionsetup{width=1\textwidth}
		\includegraphics[width=\textwidth]{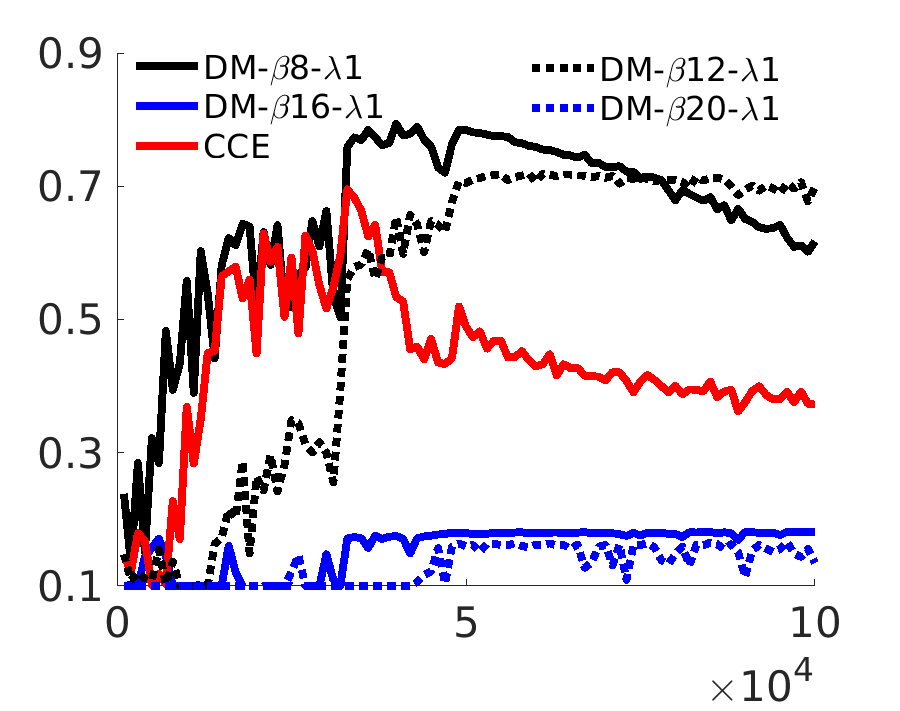}
		\label{fig:training_accuracies}
		%
	\end{subfigure}%
	\begin{subfigure}[t!]{0.245\textwidth}
		\centering
		\captionsetup{width=1\textwidth}
		\includegraphics[width=\textwidth]{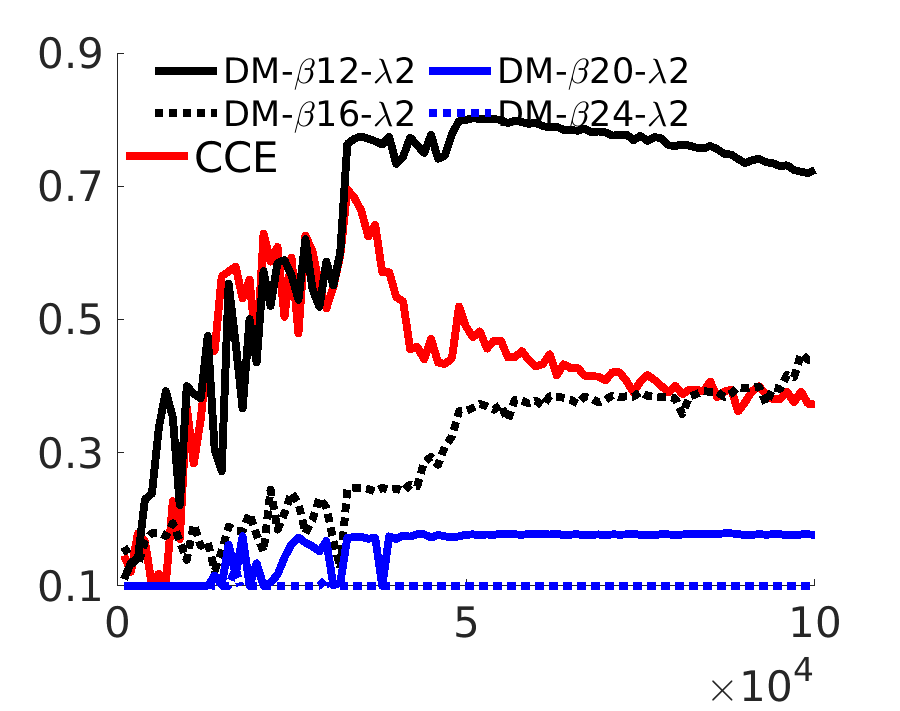}
		\label{fig:training_accuracies}
		%
	\end{subfigure}
	\caption{ResNet-56 on CIFAR-10 ($r=60\%$).
		From left to right, the results of four emphasis modes 0, $\frac{1}{3}$, 0.5, $\frac{2}{3}$ with different emphasis variances are displayed in each column respectively. 
		When $\lambda$ is larger, $\beta$ should be larger.
		Specifically :\\
		1) when $\lambda=0$: we tried $\beta=0.5, 1, 2, 4$;\\ 
		2) when $\lambda=0.5$: we tried $\beta=4, 8, 12, 16$;\\ 
		3) when $\lambda=1$: we tried $\beta=8, 12, 16, 20$; \\
		4) when $\lambda=2$: we tried $\beta=12, 16, 20, 24$.  
	}
	\label{fig:ResNet56_N06}
\end{figure*}

\begin{figure*}[h!]
	\centering
	\begin{subfigure}[t!]{0.245\textwidth}
		\centering
		\captionsetup{width=1\textwidth}
		\includegraphics[width=\textwidth]{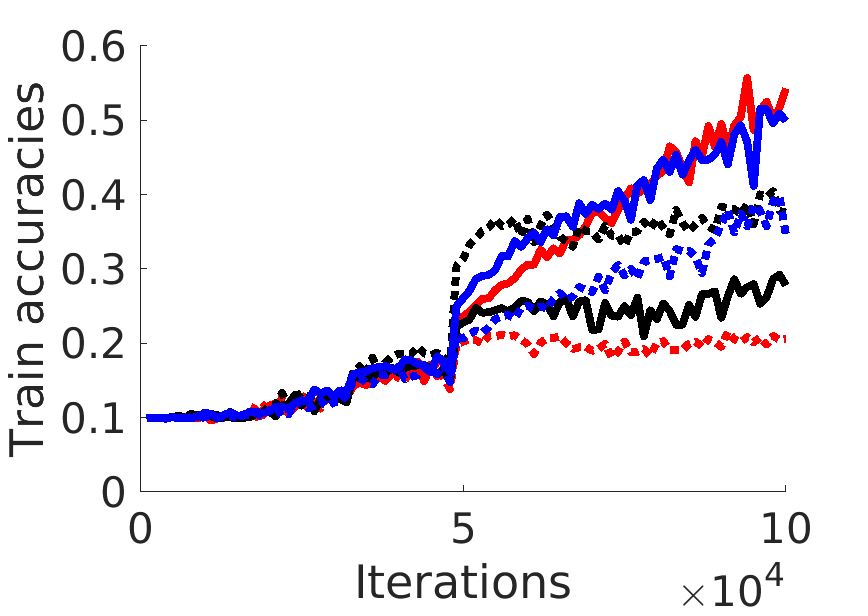}
		\label{fig:training_accuracies}
	\end{subfigure}%
	\begin{subfigure}[t!]{0.245\textwidth}
		\centering
		\captionsetup{width=1\textwidth}
		\includegraphics[width=\textwidth]{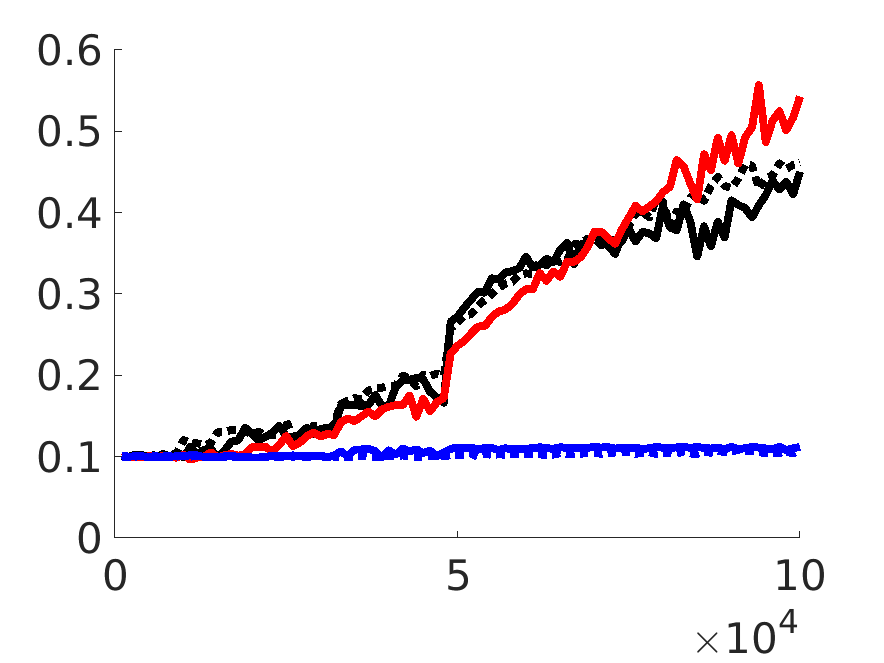}
		\label{fig:training_accuracies}
	\end{subfigure}
	\begin{subfigure}[t!]{0.245\textwidth}
		\centering
		\captionsetup{width=1\textwidth}
		\includegraphics[width=\textwidth]{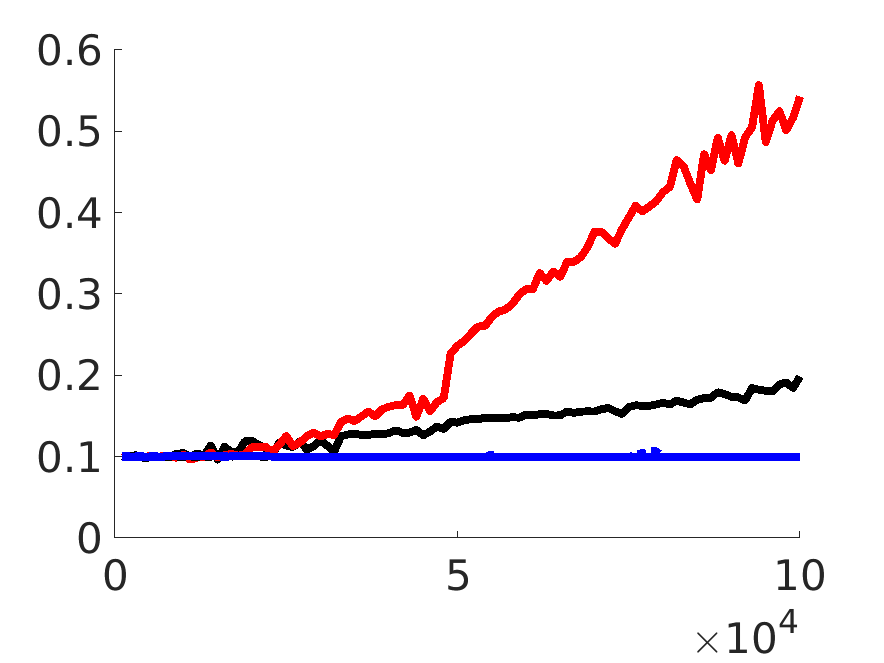}
		\label{fig:training_accuracies}
	\end{subfigure}%
	\begin{subfigure}[t!]{0.245\textwidth}
		\centering
		\captionsetup{width=1\textwidth}
		\includegraphics[width=\textwidth]{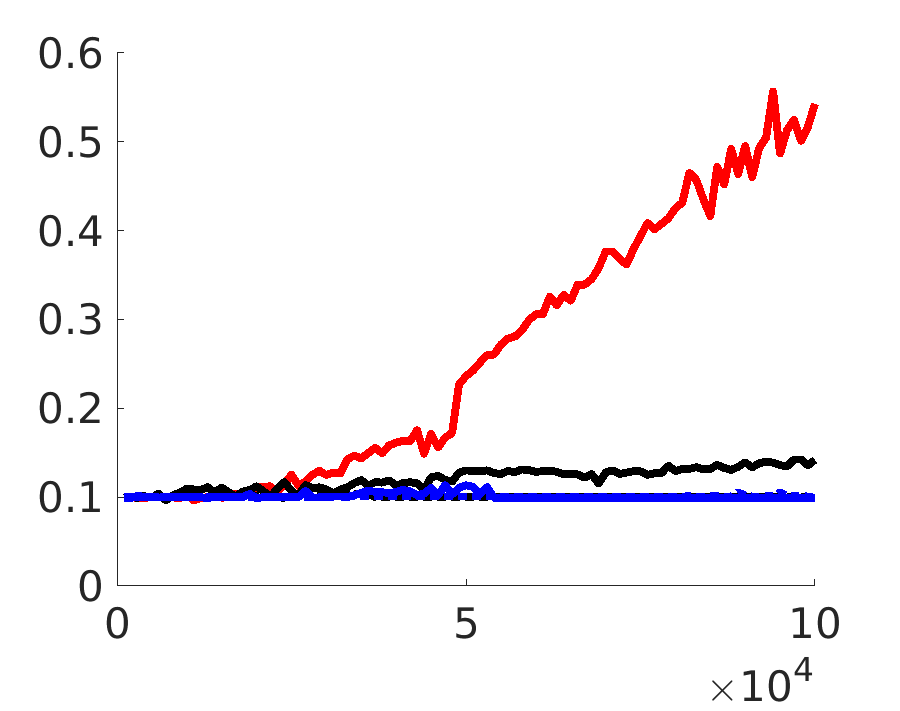}
		\label{fig:training_accuracies}
	\end{subfigure}

	\begin{subfigure}[t!]{0.245\textwidth}
		\centering
		\captionsetup{width=1\textwidth}
		\includegraphics[width=\textwidth]{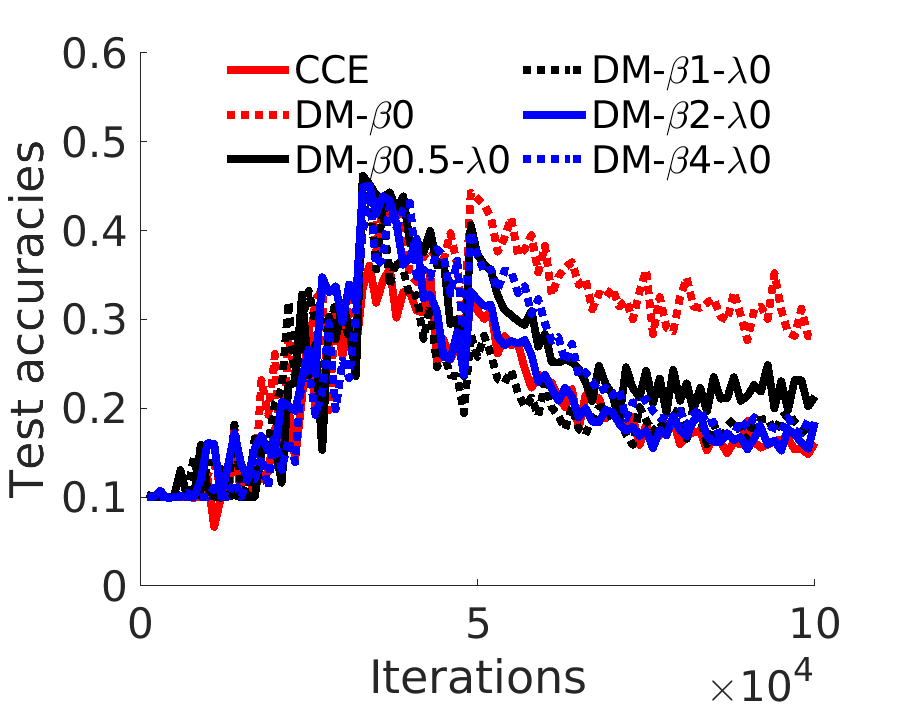}
		\label{fig:training_accuracies}
		%
	\end{subfigure}%
	\begin{subfigure}[t!]{0.245\textwidth}
		\centering
		\captionsetup{width=1\textwidth}
		\includegraphics[width=\textwidth]{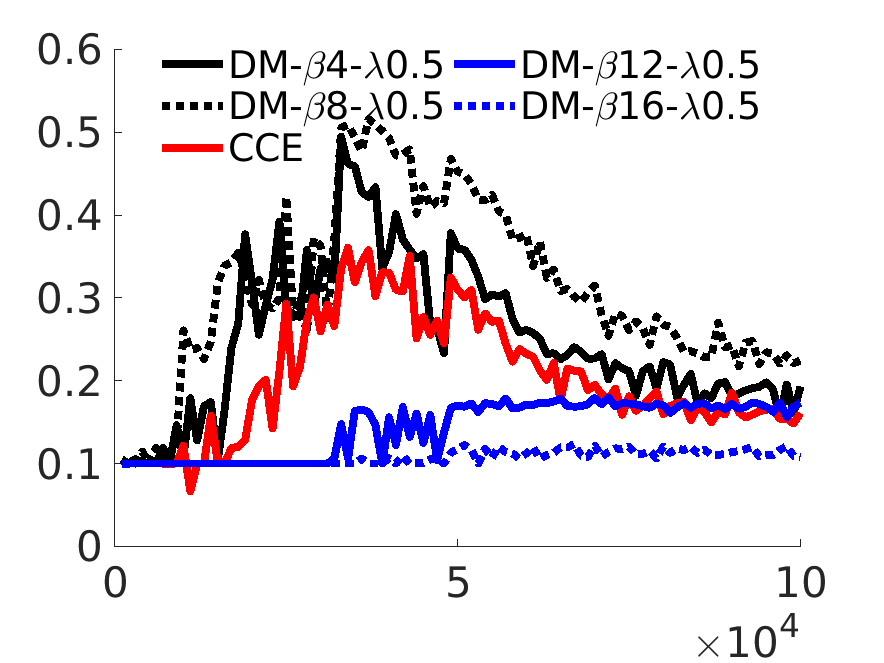}
		\label{fig:training_accuracies}
		%
	\end{subfigure}
	\begin{subfigure}[t!]{0.245\textwidth}
		\centering
		\captionsetup{width=1\textwidth}
		\includegraphics[width=\textwidth]{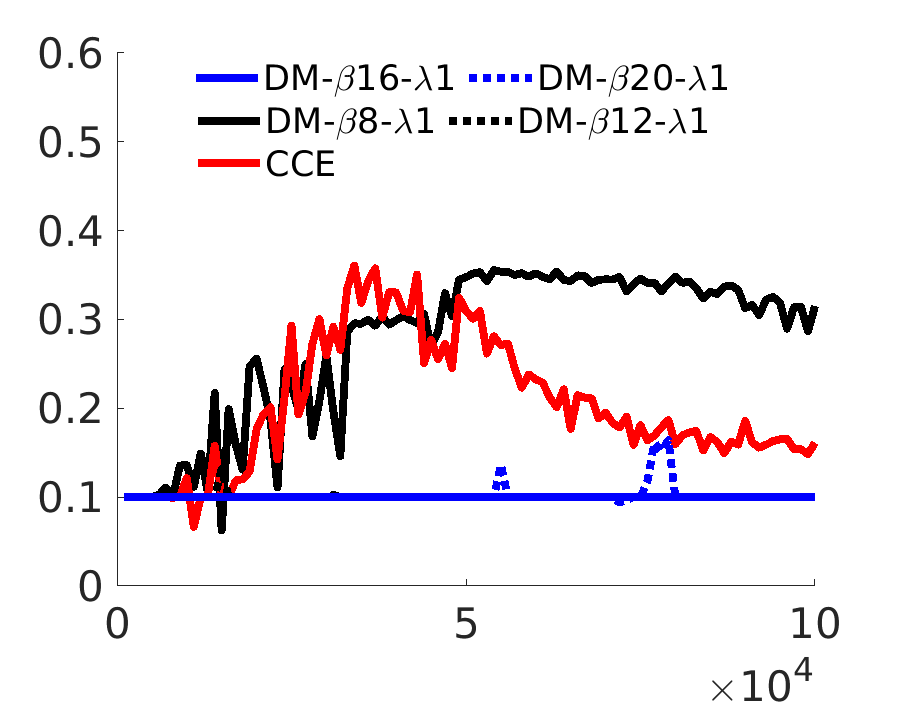}
		\label{fig:training_accuracies}
		%
	\end{subfigure}%
	\begin{subfigure}[t!]{0.245\textwidth}
		\centering
		\captionsetup{width=1\textwidth}
		\includegraphics[width=\textwidth]{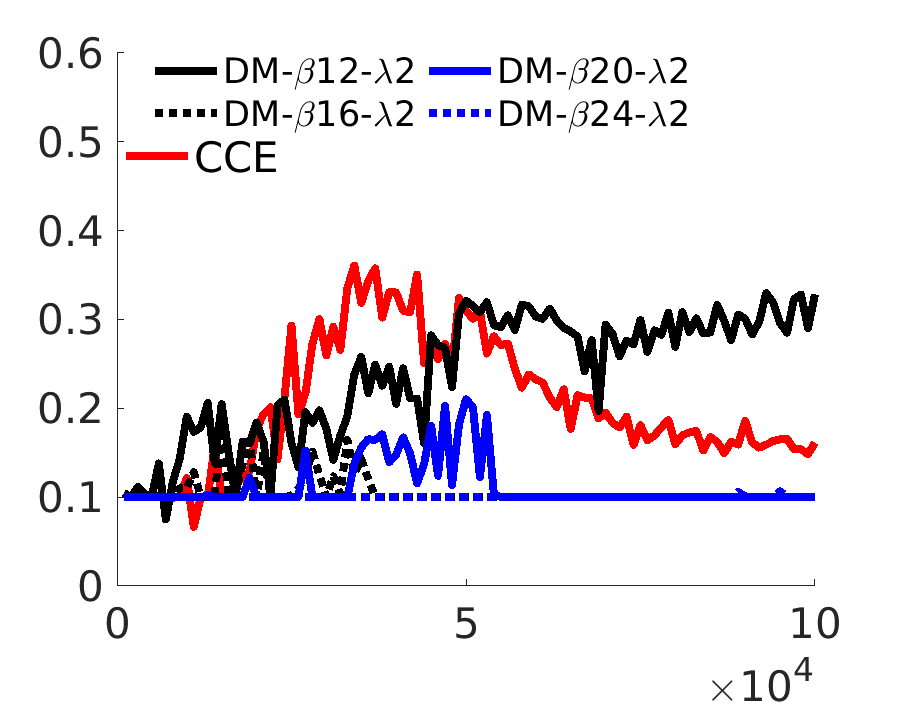}
		\label{fig:training_accuracies}
		%
	\end{subfigure}
	\caption{ResNet-56 on CIFAR-10 ($r=80\%$).
		From left to right, the results of four emphasis modes 0, $\frac{1}{3}$, 0.5, $\frac{2}{3}$ with different emphasis variances are displayed in each column respectively. 
		When $\lambda$ is larger, $\beta$ should be larger.
		Specifically :\\
		1) when $\lambda=0$: we tried $\beta=0.5, 1, 2, 4$;\\ 
		2) when $\lambda=0.5$: we tried $\beta=4, 8, 12, 16$;\\ 
		3) when $\lambda=1$: we tried $\beta=8, 12, 16, 20$; \\
		4) when $\lambda=2$: we tried $\beta=12, 16, 20, 24$.  
	}
	\label{fig:ResNet56_N08}
\end{figure*}

\end{document}